\documentclass{article}
\usepackage{blitz_preprint,times}
\usepackage{amsmath,amssymb,booktabs,graphicx,tabularx,xcolor}
\usepackage{algorithm}
\usepackage{algpseudocode}
\usepackage{hyperref}
\usepackage{placeins}
\usepackage{longtable}
\usepackage{caption}
\usepackage[letterpaper,margin=1in,headheight=14pt,headsep=18pt,footskip=30pt]{geometry}
\usepackage{etoolbox}
\AtBeginEnvironment{table}{\fontsize{10}{12}\selectfont}
\AtBeginEnvironment{thebibliography}{\fontsize{10}{12}\selectfont}
\DeclareCaptionFont{preprintcaption}{\fontsize{10}{12}\selectfont}
\renewcommand{\topfraction}{0.95}
\renewcommand{\bottomfraction}{0.9}
\renewcommand{\textfraction}{0.05}
\renewcommand{\floatpagefraction}{0.85}
\makeatletter
\renewcommand{\normalsize}{\@setfontsize\normalsize{11pt}{13pt}}
\normalsize
\DeclareMathSizes{11}{11}{8}{6}
\renewcommand{\@maketitle}{\vbox{\hsize\textwidth
\centering
{\fontsize{18}{22}\selectfont\bfseries\@title\par}
\vspace{12pt}
{\@author\par}
\vspace{15pt}}
}
\makeatother
\renewcommand{\headrulewidth}{0pt}
\renewenvironment{abstract}{\par\smallskip
\centerline{\normalsize\bfseries Abstract}\vspace{4pt}
\begin{quote}\normalsize}{\end{quote}\vspace{6pt}}
\hypersetup{hidelinks,pdfauthor={Vinay Saji Mathew, Soundar Kumara, Gretta Kellogg, William K. M. Lai},pdftitle={Blitz: Accelerating the Diffusion Modules of AlphaFold3 and Boltz-2}}
\graphicspath{{figures/}}
\newcommand{\method}{Blitz}

\title{Blitz: Accelerating the Diffusion Modules\\of AlphaFold3 and Boltz-2}
\author{%
{\fontsize{11}{13}\selectfont
Vinay Saji Mathew\textsuperscript{1}\quad
Soundar Kumara\textsuperscript{1}\quad
Gretta Kellogg\textsuperscript{2}\quad
William K. M. Lai\textsuperscript{3}}\\[7pt]
{\fontsize{9}{11}\selectfont
\textsuperscript{1}Department of Industrial and Manufacturing Engineering, The Pennsylvania State University\\
\textsuperscript{2}Institute for Computational and Data Sciences, The Pennsylvania State University\\
\textsuperscript{3}Department of Biochemistry, University at Buffalo, State University of New York}
}

\begin{document}
\maketitle
\begin{abstract}
Diffusion-based protein structure prediction models produce all-atom structures in about 200 denoising steps, and that cost has driven a number of few-step samplers. We present Blitz, a few-step distillation framework that uses a geometric noise schedule to distribute steps more evenly across noise scales, then learns those updates through multistep consistency training. Physical steering and target-free candidate selection help reduce the remaining geometric defects at inference. We apply Blitz to AlphaFold 3 and Boltz-2, reducing denoising from 200 steps to eight. Because published few-step results use different benchmarks and scoring procedures, we compare both distilled models with their teachers and other few-step methods under a common evaluation protocol covering structural accuracy, interface quality, ligand placement, and stereochemical validity. Blitz retains teacher-level structural accuracy, with AF3-Blitz outperforming all evaluated few-step baselines and exceeding its teacher on several accuracy metrics.
\end{abstract}

\section{Introduction}

Diffusion-based atom prediction has been a major breakthrough in the
biological sciences, with models such as AlphaFold3 \citep{abramson2024af3},
Boltz-1 and Boltz-2 \citep{wohlwend2024boltz1,passaro2025boltz2},
Chai-1 \citep{chai2024chai1}, and Protenix \citep{chen2025protenix}.
It builds on progress in protein folding that led CASP organizers to describe
AlphaFold2 as a solution to the decades-old folding problem for single
protein chains \citep{abbass2025casp13}. However, these models are expensive in
pipelines that require tens of thousands of predictions, leading to a
growing literature on reducing their cost. These efforts primarily tackle
three parts of the pipeline:
\begin{enumerate}
\item Reducing the cost of constructing multiple sequence alignments
(MSAs). MMseqs2 and ColabFold accelerate sequence search
\citep{steinegger2017mmseqs2,mirdita2022colabfold}, while
\citet{kallenborn2025mmseqsgpu} move the search onto GPUs.
\item Reducing the number of recycle updates and the computational cost
of the Pairformer trunk \citep{gong2025protenixmini,zhang2026dcfold}.
\item Reducing the cost of the diffusion model itself, which typically
uses 200 denoising steps in AlphaFold3 and Boltz-2
\citep{abramson2024af3,passaro2025boltz2}.
\end{enumerate}

On the model side, these costs have been approached in several ways.
\citet{gong2025protenixmini} introduced Protenix-Mini, reducing both the
cost of the trunk and the number of diffusion sampling steps.
\citet{zhang2026dcfold} introduced DCFold, whose Temporal Geodesic Matching
schedule pairs diffusion times during consistency training to improve
stability. They also apply consistency training to the Pairformer to
reduce recycling. \citet{scarpellini2026decaf} introduced DeCAF, which
distills all-atom flow maps in the noise coordinate used by EDM and applies
reward-guided search at inference. They report strong results in both
protein--ligand pose accuracy and physical validity.

However, we show here that differences in scoring conventions and
benchmark construction make these algorithms difficult to compare.
Claims of preserved quality do not fully hold under our unified evaluation. This is especially important
for few-step samplers, where steric clashes, distorted bonds, and invalid
bond angles can coexist with high structural similarity scores. Our
matched evaluations show that commonly reported accuracy metrics can
obscure these flaws when stereochemical checks are omitted
(Table~\ref{tab:conventions} and Figure~\ref{fig:structure}).

To address these gaps in stereochemical quality and structural accuracy,
we present \method{}, a few-step distillation framework. We first show
that the default polynomial noise schedule of \citet{karras2022edm} is ill-suited
to the few-step molecular sampling regime studied here, and replace it
with a geometric schedule. We then address the remaining accuracy gap
with multistep consistency training on this grid, adapting the method of
\citet{heek2024mscd}. The eight-step trajectory also leaves room for
inference-time steering, as used in Boltz-2 \citep{passaro2025boltz2}.
We tune the steering strengths for Boltz-2-Blitz and bring its
reference-free validity closer to that of its teacher. Finally, we
benchmark structural accuracy and stereochemical quality on three
popular datasets used in this literature: the Boltz-2 benchmark ($\sim$2,300 targets), Runs~N'~Poses, and RecentPDB384
\citep{passaro2025boltz2,runsnposes,chen2025protenix}. Among the compared
configurations, AF3-Blitz achieves the highest stereochecked complex lDDT
on the primary benchmark. Both students retain teacher-level structural
accuracy, exceeding their teachers on several reported metrics. We apply the same framework to AlphaFold3
and Boltz-2 to show generality across two publicly available
diffusion-based structure prediction models.

In summary, our core contributions are:
\begin{enumerate}
\item \textbf{A diagnosis.} We show that prevailing scoring and benchmark
conventions systematically mask the geometric failures of few-step
structure predictors, and we expose these failures under matched
evaluation.
\item \textbf{A method.} We introduce \method{}, which pairs a geometric
noise schedule and multistep consistency training with inference-time
physical controls to reach teacher-level structural accuracy in eight
sampling steps, a $25\times$ reduction in sampling steps for both
AlphaFold3 and Boltz-2.
\item \textbf{A protocol.} We establish a standardized, multi-metric
evaluation across three benchmarks spanning structural accuracy,
interface quality, ligand placement, and stereochemistry, designed to
surface the failure modes that single-metric reporting hides.
\end{enumerate}

\section{Background: diffusion and consistency distillation}
\label{sec:background}

\paragraph{Diffusion in AlphaFold3.}
AF3 adopts the EDM formulation and polynomial noise schedule of
\citet{karras2022edm} for all-atom structure prediction
\citep[Supplementary Information, Sections 3.7--3.7.1]{abramson2024af3}.
Its diffusion module is a non-equivariant transformer conditioned on the
single and pair representations from the trunk. Let
$x_0\in\mathbb R^{N\times3}$ contain the resolved atomic coordinates and
$c$ the molecular features and trunk conditioning. In these
variance-exploding coordinates, a training input is
\begin{equation}
 x_\sigma=x_0+\sigma\epsilon,\qquad \epsilon\sim\mathcal N(0,I).
 \label{eq:noise}
\end{equation}
The denoiser $D_\phi(x_\sigma,\sigma;c)$ predicts clean coordinates.
AF3 uses EDM input/output preconditioning and samples training noise as
$\sigma=\sigma_{\mathrm{data}}\exp(-1.2+1.5z)$, with
$z\sim\mathcal N(0,1)$ and $\sigma_{\mathrm{data}}=16$~\AA{}.
The trunk is evaluated once and reused across multiple noised copies;
the coordinate loss uses weighted rigid alignment.

\paragraph{The EDM noise schedule.}
Sampling begins at high noise and follows a decreasing sequence of noise
levels. Karras et al.\ space these levels polynomially. For $K\geq2$
positive nodes, we write
\begin{equation}
 \sigma_i=\left[\sigma_{\max}^{1/\rho}
 +\frac{i}{K-1}\left(\sigma_{\min}^{1/\rho}
 -\sigma_{\max}^{1/\rho}\right)\right]^\rho,
 \qquad i=0,\ldots,K-1.
 \label{eq:karras}
\end{equation}
AF3 uses $\rho=7$, with absolute noise endpoints
$\sigma_{\max}=2560$~\AA{} and $\sigma_{\min}=0.0064$~\AA{}
\citep[Supplementary Eq.\ 7]{abramson2024af3}.
EDM selected $\rho=7$ through experiments on image generation.

\paragraph{Consistency models.}
Consistency models learn to map points on the same probability-flow
trajectory to its low-noise endpoint \citep{song2023consistency}.
For two states $x_s,x_t$ on that trajectory, the desired property is
$f_\theta(x_s,s)=f_\theta(x_t,t)$, with the boundary condition
$f_\theta(x_{\varepsilon},\varepsilon)=x_{\varepsilon}$ at a small
positive endpoint $\varepsilon$.
In consistency distillation, a frozen diffusion teacher takes a numerical
step to form a pair of nearby states. Training brings their student
predictions together, stopping gradients through the target branch.
Consistency training instead forms pairs directly from noised data,
without a pretrained teacher \citep{song2023consistency,song2024ict}.

\paragraph{Multistep consistency models.}
\citet{heek2024mscd} unify TRACT \citep{berthelot2023tract} and
consistency models \citep{song2023consistency} by dividing the trajectory
into fixed segments. A shared model maps each point to the lower
boundary of its segment.
Generation traverses the segments in order, using one evaluation per
segment. Their formulation uses variance-preserving coordinates,
\begin{equation}
 z_t=\alpha_t x_0+\beta_t\epsilon,\qquad
 \alpha_t^2+\beta_t^2=1,
 \label{eq:vp}
\end{equation}
and constructs the segment maps with DDIM updates.

\section{What does it mean to preserve quality in structure prediction?}
\label{sec:quality}
\begin{figure}[!htb]
\centering\includegraphics[width=.86\linewidth]{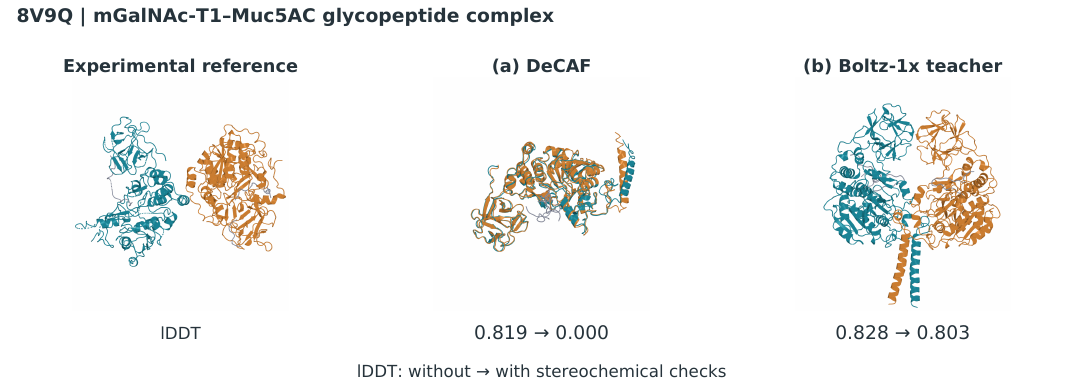}
\caption{Selected 8V9Q example \citep{collette2024galnact1}: experimental
reference, (a)~released DeCAF, and (b)~its Boltz-1x teacher.
Scores show lDDT without $\to$ with stereochemical checks.}
\label{fig:structure}
\end{figure}

The prediction in Figure~\ref{fig:structure}a has severe overlap between
its protein chains. Despite having recognizable individual folds, this
arrangement is sterically implausible; however, it still scores well on
unchecked lDDT. Its Boltz-1x teacher, shown in
Figure~\ref{fig:structure}b, retains recognizable folds without this
severe overlap; this raises an important question: \textit{what must
accelerated structure prediction preserve when we remove these denoising
steps?}

This is particularly important when there is a claim that a faster sampler
preserves accuracy. Unchecked lDDT measures how well local interatomic
distances are preserved from the reference. However, this local fold
preservation can coexist with severe clashes and violations of bond
lengths and angles. The original lDDT authors anticipated this distinction
and provided optional stereochemical checks to account for these
violations \citep{mariani2013lddt}. PoseBusters showed that agreement with an experimental ligand pose does not
necessarily establish physical plausibility \citep{buttenschoen2024posebusters}.
However, evaluations like those done by DCFold report only RMSD thresholds
on the PoseBusters benchmark without reporting PB-validity, leaving the
failure mode that motivated those checks unexamined \citep{zhang2026dcfold}. This concern also applies to cofolding.
AlphaFold3 documents overlapping chains \citep{abramson2024af3}, and DeCAF
reports ligand validity alongside pose accuracy
\citep{scarpellini2026decaf}. Figure~\ref{fig:structure} helps illustrate
the discrepancy at the protein-assembly level.

Table~\ref{tab:conventions} helps make this discrepancy in the evaluated
configurations more explicit. On matched targets from the Boltz-2 benchmark ($\sim$2,300 targets),
the default Protenix sampler and its one-step ODE counterpart are almost
identical on unchecked lDDT. DCFold uses a corresponding Protenix-based
one-step ODE configuration as its untrained AF3 ODE baseline
\citep{zhang2026dcfold}. However, with stereochemical checks enabled, our
one-step configuration's score falls approximately 32 points on a
100-point scale, while the default sampler's score changes only slightly.
Furthermore, the one-step configuration has, on average, approximately
1,500 counted bond-length, bond-angle, and clash violations per target.

\begin{table}[!htb]
\centering
\caption{\textbf{Accuracy with and without stereochemical checks.}
Bold highlights geometry failures. NFE counts steering particles per prediction.
Evaluation details are in Appendix~\ref{app:evaluation}.}
\label{tab:conventions}
\setlength{\tabcolsep}{3pt}
\begin{tabular}{lrrrrrrr}
\toprule
& & PXM $\uparrow$ & \multicolumn{2}{c}{OST lDDT $\uparrow$} & Events $\downarrow$ & & PB-valid $\uparrow$ \\
\cmidrule(lr){4-5}
Sampler & NFE & lDDT & Unchecked & Checked & /target & lDDT-PLI $\uparrow$ & valid \\
\midrule
AF3 teacher & 200 & 0.8554 & 0.8626 & 0.8615 & 0.7 & 0.6446 & 44.8\% \\
AF3-Blitz [8] & 8 & 0.8579 & 0.8653 & 0.8635 & 4.8 & 0.6430 & 42.7\% \\

Protenix & 200 & 0.8512 & 0.8593 & 0.8557 & 4.3 & 0.6259 & 45.8\% \\
Protenix one-step ODE & 1 & 0.8503 & \textbf{0.8591} & \textbf{0.5426} & \textbf{1563.6} & 0.5774 & \textbf{26.0\%} \\

Boltz-2x teacher & 600 & 0.8426 & 0.8514 & 0.8499 & 4.4 & 0.5596 & 60.9\% \\
Boltz-2-Blitz [8] & 8 & 0.8451 & 0.8542 & 0.8492 & 14.2 & 0.5545 & 52.0\% \\

Boltz-1x teacher & 600 & 0.8272 & 0.8357 & 0.8074 & 35.8 & 0.5932 & 58.0\% \\
DeCAF default & 10 & 0.8254 & \textbf{0.8340} & \textbf{0.6756} & \textbf{681.3} & 0.5872 & \textbf{25.0\%} \\
DeCAF SDE ($\gamma=0.5$) & 10 & 0.8166 & \textbf{0.8250} & \textbf{0.6635} & \textbf{680.6} & 0.5810 & \textbf{26.6\%} \\
\bottomrule
\end{tabular}
\par\smallskip\parbox{\linewidth}{PXM: Protenix's PXMeter lDDT scorer. OST: OpenStructure's
\texttt{compare-structures} scorer, which enables stereochemical checks
by default; ``Unchecked'' disables them.}
\end{table}

This gap is one part of the picture. DockQ, PoseBusters validity, ICS,
IPS, and lDDT-PLI provide complementary information alongside
stereochemically checked lDDT. In essence, we argue that these
measures should be reported where applicable, with their definitions and
evaluation settings made explicit, particularly for fast samplers to
provide an accurate picture. Appendix~\ref{app:evaluation} gives our
scoring rules and check coverage.

\section{Blitz}
\label{sec:method}

\subsection{Noise allocation before distillation}
\label{sec:numerical}
\begin{figure}[t]
\centering\includegraphics[width=.86\linewidth]{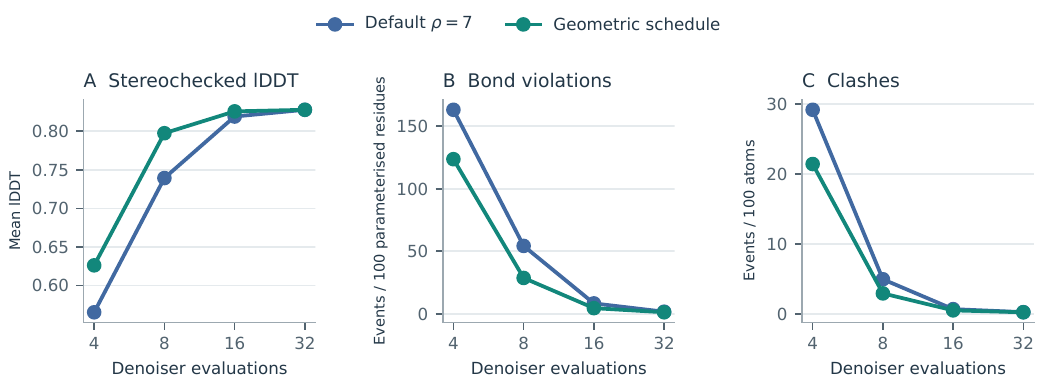}
\caption{\textbf{Allocation matters most at low budgets.} The same
untrained AF3 denoiser is evaluated with the default and geometric
schedules at matched step budgets and positive endpoints.}
\label{fig:budget}
\end{figure}

Before detailing our distillation procedure, we identify a problem with
the noise schedule in the low-step regime. Following the deterministic
analysis of \citet{karras2022edm}, we hold the pretrained AF3 denoiser
and solver fixed and replace the default Karras schedule with a geometric
schedule. Stereochemically checked lDDT improves while bond violations
and clashes decrease, as shown in Figure~\ref{fig:budget}. Critically,
the improvement is largest at the smallest step budgets. The same pattern holds across every
publicly available structure generation model we evaluated.

The geometric schedule is the $\rho\to\infty$ limit of
Equation~\ref{eq:karras}:
\begin{equation}
 \sigma_i=\sigma_{\max}^{1-u_i}\sigma_{\min}^{u_i},\qquad
 u_i=\frac{i}{K-1},\quad i=0,\ldots,K-1.
 \label{eq:geometric-limit}
\end{equation}
It spaces the positive nodes uniformly in log noise, giving the same
relative noise reduction between adjacent nodes. In Section~\ref{sec:results},
we compare students trained identically except for the noise schedule:
polynomial with $\rho=7$ or geometric.

\subsection{Local approximation error and endpoint damage}
\label{sec:local-error}
We next ask: \textit{Why does changing the schedule improve molecular
geometry?} With fewer steps, each update must cover a larger noise
interval. We first measure what is lost by crossing that interval in
one step. On selected diagnostic targets, we start at the same state
$x_i$ and compare one Euler step $E_i$ with a Heun calculation
$\widehat\Phi_i^{(m)}$ that divides the interval into $m$ smaller steps:
\begin{equation}
 e_i^{(m)}=E_i(x_i)-\widehat\Phi_i^{(m)}(x_i),\qquad
 \varepsilon_i^{(m)}=\frac{\|e_i^{(m)}\|_F}{\sqrt{3N}}.
 \label{eq:local}
\end{equation}
This estimates the coordinate error introduced by the coarse step,
measured in the same frame over $N$ valid atoms.

The next question is: \textit{Does this error damage the final molecule?}
We continue both endpoints with the same refined sampler $R_i^{(m)}$,
including the same terminal step, and measure the difference in final
quality:
\begin{equation}
 \Delta q_i^{(m)}=
 q\!\left(R_i^{(m)}(E_i(x_i))\right)
 -q\!\left(R_i^{(m)}(\widehat\Phi_i^{(m)}(x_i))\right).
 \label{eq:effect}
\end{equation}
Here $q$ is stereochemically checked lDDT or a bond, angle, or clash
violation count. Changing only one interval isolates its effect on the
final structure.

\begin{figure}[t]
\centering\includegraphics[width=.86\linewidth]{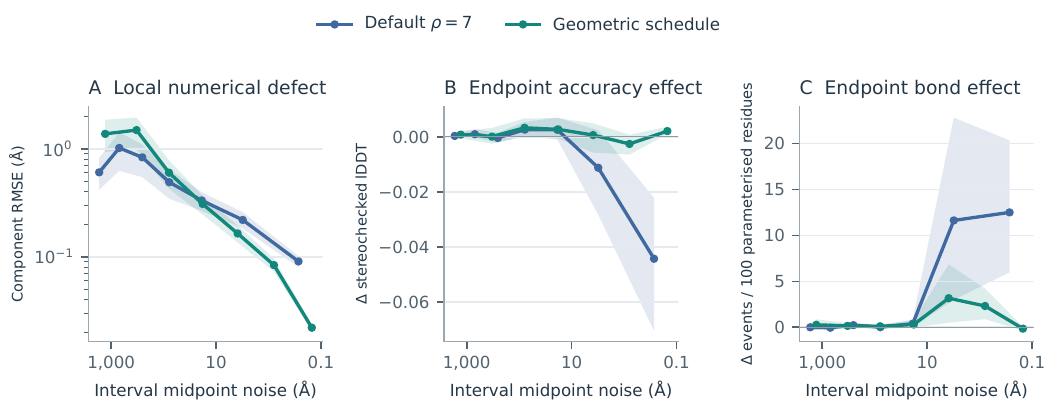}
\caption{\textbf{Local approximation error and final molecular quality.}
A: Euler error relative to a refined Heun calculation.
B--C: final effects of replacing one refined interval with an Euler step.
Details are in
Appendix~\ref{app:numerical}.}
\label{fig:interval}
\end{figure}

Indeed, the final positive interval of the default schedule causes the
largest average loss of stereochemically checked lDDT, even though an
earlier interval causes a larger final coordinate displacement
(Figure~\ref{fig:interval}). Coordinate error alone therefore does not
explain the damage to molecular geometry. This gives us a reason to
redistribute the noise intervals before distillation, as the geometric
schedule does. Appendix~\ref{app:numerical} gives the full analysis and
checks on the numerical reference.

\subsection{Multistep consistency distillation}
We adapt multistep consistency distillation \citep{heek2024mscd} to the
variance-exploding EDM coordinates used by AF3 and Boltz-2. The geometric
grid divides the trajectory into $K$ fixed segments, with boundaries
$b_0>\cdots>b_K=0$. We refine a separate teacher grid during training,
keeping these segment boundaries fixed. We sample an adjacent edge using
EDM log-normal importance weights, then identify its student segment.

DCFold reports instability with conventional consistency training
\citep{zhang2026dcfold}, while DeCAF finds poor physical validity with
its consistency-distillation baseline \citep{scarpellini2026decaf}.
Relaxing the one-step requirement shortens the mapping the student must
learn. In our setting, multistep distillation works with an aligned
coordinate-space loss and ordinary EDM edge sampling, without DCFold's
Temporal Geodesic Matching.

A short Heun step of the frozen teacher is followed by a stopped-gradient
student update to the segment boundary. Inverting
Equation~\ref{eq:deployment} with $S=1$ gives the denoiser target $y^*$
in Algorithm~\ref{alg:training}. Here $c$ denotes molecular input features,
$x_0$ the reference coordinates, and $a$ the atom weights, zero for unresolved
or padded atoms. The conditioning $h$ includes $c$ and the frozen trunk
embeddings. We apply weighted Kabsch alignment and a pseudo-Huber loss;
Appendix~\ref{app:training} defines the loss and boundary updates.

\begin{algorithm}[!htb]
\caption{\method{} multistep consistency distillation}
\label{alg:training}
\begin{algorithmic}
\Statex \textbf{Input:} training set $\mathcal D$, frozen trunk $H$, frozen
teacher $D_{\phi}$, student $D_{\theta}$ initialized from $D_{\phi}$,
segment boundaries $b_0>\cdots>b_K=0$
\For{each optimizer update}
    \State Sample $(c,x_0,a)\sim\mathcal D$, augment $x_0$, and set $h\gets(c,H(c))$
    \State Sample an edge $(s,s^-)$ from the current teacher grid
    \State Set $b\gets b_{j+1}$ for the segment containing $s$, and $s^-\gets\max(s^-,b)$
    \State Sample $\epsilon\sim\mathcal N(0,I)$ and set $x_s\gets x_0+s\epsilon$
    \State $x_{s^-}\gets\Call{TeacherStep}{D_{\phi},x_s,s,s^-,h}$
    \State $x_b^{\mathrm{ref}}\gets x_{s^-}$ if $s^-=b$; otherwise
    $x_b^{\mathrm{ref}}\gets\Call{Step}{D_{\theta},x_{s^-},s^-,b,h}$
    \State $y^*\gets\operatorname{stopgrad}\!\left[(sx_b^{\mathrm{ref}}-b x_s)/(s-b)\right]$
    \State $\hat y_{\theta}\gets\operatorname{Align}_a(D_{\theta}(x_s,s,h),y^*)$
    \State $\mathcal L\gets w(s)\operatorname{PH}_{a,\delta}(\hat y_{\theta}-y^*)$
    \State Average $\nabla_{\theta}\mathcal L$ over noised copies and training examples; update $\theta$
\EndFor
\end{algorithmic}
\end{algorithm}

The student is initialized from its teacher; only the diffusion module
is trained. Each trunk evaluation conditions multiple noised copies.
At inference, Equation~\ref{eq:deployment} composes the $K$
learned maps. Appendix~\ref{app:training} gives training parameters,
boundary handling, and deployment scales.

\subsection{Physical steering and target-free selection}
The Boltz-2 students use physical steering during sampling, with
strengths approximated from the teacher's coordinate corrections. A correction $c_i$ to the denoised estimate changes the
next state by $\alpha_i c_i$, where $r_i=\sigma_{i+1}/\sigma_i$ and
\begin{equation}
 \alpha_i=S(1-r_i)\quad\text{(Euler)},\qquad
 \alpha_i=1-r_i^S\quad\text{(exponential update)}.
 \label{eq:steering-transfer}
\end{equation}
For each potential, we sum correction vectors before taking their RMS
and compare transferred corrections in matched log-noise bins. This
guides the weights, which we then validate using accuracy and geometry.
This steering did not work well for AF3 in our tests.

Blitz reranks five candidates without reference structures or further
 diffusion calls. We prefer its lower geometry-violation rate;
Appendix~\ref{app:native-selection} compares native confidence.

\raggedbottom
\section{Data and evaluation}
\label{sec:results}

We evaluate Blitz using AlphaFold3 \citep{abramson2024af3} and Boltz-2 \citep{passaro2025boltz2} to assess generality across two all-atom structure prediction systems. The schedule improvements in Sections~\ref{sec:numerical}--\ref{sec:local-error} also hold for Chai-1 and Protenix \citep{chai2024chai1,chen2025protenix}.

\paragraph{Benchmarks and reporting.}
We evaluate the Boltz-2 benchmark ($\sim$2,300 targets), Runs~N'~Poses, and
Protenix's 384-target recent-PDB sample
\citep{passaro2025boltz2,runsnposes,chen2025protenix}. Main configurations
use three recycles and five candidates.
For brevity, the main text focuses on the Boltz-2 benchmark ($\sim$2,300 targets).
Results for Runs~N'~Poses and RecentPDB384, including ligand-evaluation
coverage limitations, are reported in Appendices~\ref{app:rnp} and~\ref{app:recentpdb}.

\paragraph{Accuracy and physical quality.}
We use OpenStructure \citep{biasini2013openstructure} to score complex
and backbone lDDT, interfaces, ligand placement, and bond, angle, and
clash violations. Complex lDDT includes stereochemical checks
\citep{mariani2013lddt}. Ligand validity is evaluated separately with
DeCAF-compatible PoseBusters \citep{buttenschoen2024posebusters}.
Success requires the same primary ligand to pass PB-valid and RMSD
$<2$~\AA{} in the same candidate.
We also report the fraction of targets with geometry defects. Failed and inapplicable evaluations remain separate from invalid predictions.
Appendix~\ref{app:uncertainty} reports confidence intervals for mean scores
and paired teacher--student differences.

\begin{table}[!htb]
\centering
\caption{\textbf{Accuracy and geometry across sampling budgets.}}
\label{tab:af3}\label{tab:parity}
\setlength{\tabcolsep}{2.5pt}
\setlength{\tabcolsep}{3pt}\begin{tabular}{@{}lrrrrrrrr@{}}\toprule
& \multicolumn{5}{c}{AlphaFold3} & \multicolumn{3}{c}{Boltz-2} \\
\cmidrule(lr){2-6}\cmidrule(lr){7-9}
Metric & Teacher & $\rho{=}7$ & \shortstack{\textbf{Blitz}\\\textbf{[4]}} & \shortstack{\textbf{Blitz}\\\textbf{[8]}} & \shortstack{\textbf{Blitz}\\\textbf{[16]}} & Teacher & \shortstack{\textbf{Blitz}\\\textbf{[8]}} & \shortstack{\textbf{Blitz}\\\textbf{[16]}} \\
\midrule
NFE & 200 & 8 & 4 & 8 & 16 & 600 & 8 & 16 \\
\midrule
Complex lDDT $\uparrow$ & 0.8615 & 0.8603 & 0.8559 & \textbf{0.8635} & \textbf{0.8634} & 0.8499 & 0.8492 & \textbf{0.8533} \\
Backbone lDDT $\uparrow$ & 0.9136 & \textbf{0.9151} & 0.9128 & \textbf{0.9150} & \textbf{0.9151} & 0.9043 & \textbf{0.9055} & \textbf{0.9069} \\
TM-score $\uparrow$ & 0.8898 & 0.8884 & 0.8889 & 0.8888 & 0.8894 & 0.8753 & \textbf{0.8768} & \textbf{0.8777} \\
DockQ $\uparrow$ & 0.4198 & 0.4133 & 0.4185 & \textbf{0.4207} & 0.4167 & 0.3929 & 0.3897 & \textbf{0.3932} \\
QS-global $\uparrow$ & 0.7784 & 0.7756 & \textbf{0.7805} & 0.7768 & 0.7784 & 0.7374 & 0.7347 & 0.7355 \\
ICS $\uparrow$ & 0.4263 & 0.4211 & \textbf{0.4288} & 0.4263 & 0.4247 & 0.3884 & 0.3857 & 0.3864 \\
IPS $\uparrow$ & 0.4560 & 0.4538 & \textbf{0.4573} & \textbf{0.4574} & \textbf{0.4570} & 0.4225 & 0.4224 & \textbf{0.4230} \\
\midrule
lDDT-PLI $\uparrow$ & 0.6437 & 0.6434 & 0.6372 & 0.6425 & 0.6437 & 0.5596 & 0.5545 & 0.5573 \\
Ligand RMSD (\AA) $\downarrow$ & 6.63 & 6.51 & \textbf{6.44} & \textbf{6.44} & \textbf{6.51} & 9.00 & \textbf{8.75} & \textbf{8.78} \\
Assigned $<2$\AA{} (\%) $\uparrow$ & 61.3 & 61.4 & 61.0 & 61.3 & \textbf{61.4} & 52.2 & 51.6 & 52.2 \\
PB-valid (\%) $\uparrow$ & 45.0 & 35.2 & 34.7 & 43.0 & \textbf{45.7} & 60.9 & 52.0 & 55.9 \\
Primary $<2$\AA{} (\%) $\uparrow$ & 68.0 & 67.3 & 67.5 & 68.0 & \textbf{68.3} & 60.9 & 60.3 & 60.7 \\
Success (\%) $\uparrow$ & 34.7 & 29.6 & 30.4 & \textbf{35.8} & \textbf{35.9} & 36.5 & 30.6 & 33.9 \\
RF-valid (\%) $\uparrow$ & 41.5 & 36.4 & 36.9 & 40.5 & \textbf{42.3} & 98.5 & 93.7 & 94.1 \\
\midrule
Bonds / target $\downarrow$ & 0.06 & 2.84 & 12.84 & 0.79 & 0.10 & 0.90 & 4.83 & 1.79 \\
Angles / target $\downarrow$ & 0.09 & 1.02 & 4.00 & 0.23 & \textbf{0.08} & 0.71 & 3.05 & 1.23 \\
Clashes / target $\downarrow$ & 0.54 & 4.03 & 24.06 & 3.80 & 1.60 & 2.82 & 6.30 & 3.63 \\
Any violation (\%) $\downarrow$ & 9.8 & 50.6 & 57.5 & 11.3 & \textbf{6.8} & 30.3 & 41.9 & \textbf{26.7} \\
\bottomrule\end{tabular}\par\smallskip\parbox{\linewidth}{NFE is per prediction; assigned and primary denote ligand RMSD. All Boltz-2 configurations use three recycle updates and steering; step scales are given in Appendix~\ref{app:contracts}. Evaluation details are in Appendix~\ref{app:evaluation}.}

\end{table}

\subsection{Accuracy and geometry with fewer steps}
\begin{figure}[!htb]
\centering\includegraphics[width=.78\linewidth]{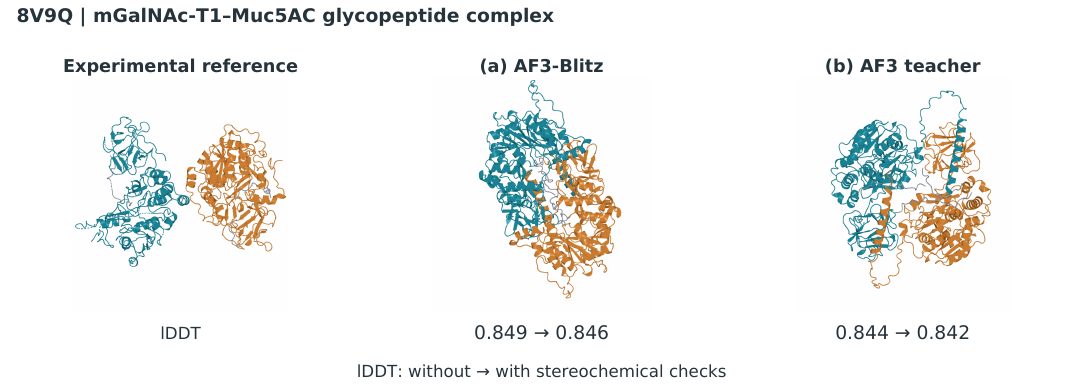}
\caption{The same 8V9Q example as Figure~\ref{fig:structure}: reference,
(a)~AF3-Blitz, and (b)~its AF3 teacher. Scores show unchecked $\to$
stereochemically checked lDDT.}
\label{fig:structure-blitz}
\end{figure}

Table~\ref{tab:af3} shows that both eight-step Blitz students retain
much of their teachers' structural accuracy with a fraction of the
sampling steps. The four- and sixteen-step AF3 students show how this
trade-off changes with the sampling budget. The composition breakdown in
Appendix~\ref{app:composition} shows higher checked lDDT for both AF3 students
on protein-only and protein--ligand complexes; gains are not uniform
across nucleic-acid-containing complexes.

\begin{samepage}
We return to the example that opened Section~\ref{sec:quality}.
Figure~\ref{fig:structure-blitz} shows that AF3-Blitz retains the individual
folds without the severe chain overlap seen in Figure~\ref{fig:structure}.
Its stereochemically checked score also remains close to its unchecked
score. This is the improvement we want from a fast sampler: preserving
the fold without the same collapse in molecular geometry.\par
\end{samepage}

\begin{table}[!htb]
\centering
\caption{\textbf{Other sampler configurations.} Metrics match Table~\ref{tab:af3}.}
\label{tab:other-samplers}
\setlength{\tabcolsep}{2.5pt}
\begin{tabular}{lrrrrrrr}\toprule
Metric & Boltz-1x & \shortstack{DeCAF\\default} & \shortstack{DeCAF\\SDE} & \shortstack{DeCAF\\FK} & Protenix & \shortstack{Protenix\\ODE} & \shortstack{Protenix\\Mini} \\
\midrule
NFE & 600 & 10 & 10 & 40 & 200 & 1 & 1 \\
\midrule
Complex lDDT $\uparrow$ & 0.8051 & 0.6741 & 0.6607 & 0.6654 & 0.8533 & 0.5462 & 0.6571 \\
Backbone lDDT $\uparrow$ & 0.8928 & 0.8869 & 0.8784 & 0.8791 & 0.9071 & 0.9039 & 0.8938 \\
TM-score $\uparrow$ & 0.8708 & 0.8622 & 0.8581 & 0.8577 & 0.8807 & 0.8780 & 0.8680 \\
DockQ $\uparrow$ & 0.3868 & 0.3702 & 0.3626 & 0.3636 & 0.4077 & 0.4026 & 0.3684 \\
QS-global $\uparrow$ & 0.7332 & 0.7247 & 0.7196 & 0.7257 & 0.7561 & 0.7483 & 0.7183 \\
ICS $\uparrow$ & 0.3842 & 0.3740 & 0.3698 & 0.3757 & 0.4071 & 0.3996 & 0.3686 \\
IPS $\uparrow$ & 0.4228 & 0.4171 & 0.4141 & 0.4177 & 0.4393 & \textbf{0.4399} & 0.4198 \\
\midrule
lDDT-PLI $\uparrow$ & 0.5934 & 0.5874 & 0.5810 & 0.5876 & 0.6244 & 0.5780 & 0.5781 \\
Ligand RMSD (\AA) $\downarrow$ & 7.01 & \textbf{6.97} & 7.11 & 7.03 & 7.00 & \textbf{6.68} & 7.46 \\
Assigned $<2$\AA{} (\%) $\uparrow$ & 57.3 & 56.0 & 56.1 & 56.5 & 59.5 & 54.9 & 54.3 \\
PB-valid (\%) $\uparrow$ & 58.2 & 24.7 & 26.7 & 53.9 & 45.8 & 26.7 & 28.2 \\
Primary $<2$\AA{} (\%) $\uparrow$ & 66.3 & 65.7 & 65.8 & 65.6 & 69.4 & 65.7 & 63.8 \\
Success (\%) $\uparrow$ & 39.4 & 21.1 & 21.8 & 38.4 & 34.0 & 25.0 & 24.0 \\
RF-valid (\%) $\uparrow$ & 99.0 & 26.0 & 27.5 & 66.8 & 32.7 & 30.9 & 31.4 \\
\midrule
Bonds / target $\downarrow$ & 11.59 & 268.56 & 271.91 & 280.02 & 0.19 & 666.44 & 373.96 \\
Angles / target $\downarrow$ & 5.45 & 123.85 & 131.80 & 138.37 & 0.58 & 283.84 & 129.20 \\
Clashes / target $\downarrow$ & 18.79 & 288.56 & 276.60 & 228.94 & 3.71 & 636.58 & 180.51 \\
Any violation (\%) $\downarrow$ & 95.0 & 97.0 & 97.5 & 97.7 & 49.1 & 100.0 & 99.95 \\
\bottomrule\end{tabular}\par\smallskip\parbox{\linewidth}{ DeCAF FK uses four steering particles; Boltz-1x uses three. Evaluation details follow Table~\ref{tab:af3}.}

\end{table}

This distinction also appears in the comparison with other few-step
samplers (Table~\ref{tab:other-samplers}). The evaluated DeCAF configurations
and Protenix one-step ODE baseline have backbone scores much closer to
Blitz than their stereochemically checked all-atom scores. The fold alone
does not capture the difference in molecular geometry.

DeCAF default uses the released ten-step sampler ($\gamma=1$); its SDE
variant uses $\gamma=0.5$ without guidance. DeCAF FK uses four particles
and ten steps, requiring 40 denoiser evaluations per retained pose.
FK steering improves ligand validity, while substantial protein-geometry
violations remain.

\subsection{What does the average hide?}
\label{sec:ablation}
\begin{figure}[!htb]
\centering\includegraphics[width=.86\linewidth]{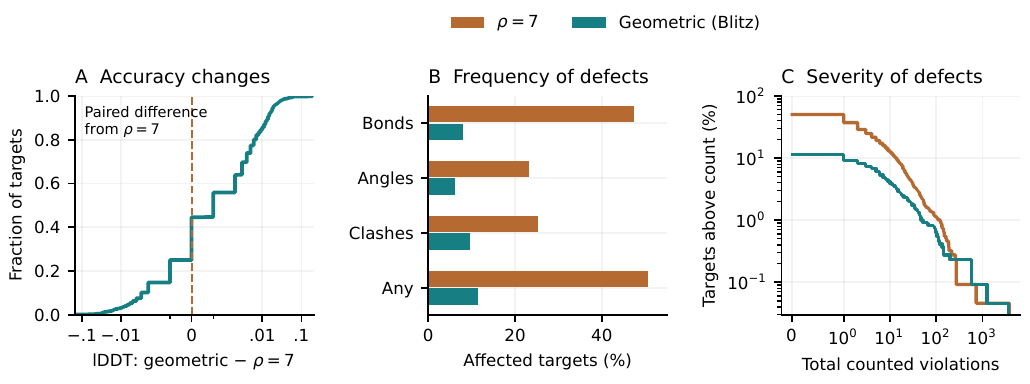}
\caption{\textbf{Similar averages, different failure patterns.}
A: paired lDDT changes (geometric minus $\rho=7$); zero denotes equal
scores. B--C: defect frequency and
violation-count tails on matched Boltz-2 benchmark targets.
A uses a signed-log change axis; C uses log-scaled counts and percentages,
with a linear region near zero count.
Zero counted violations does not imply full chemical validity.}
\label{fig:schedule-violations}
\end{figure}

The polynomial ($\rho=7$) and geometric students use the same training
setup, differing only in the noise schedule. They have close average
lDDT, but their errors are distributed very differently.
Figure~\ref{fig:schedule-violations} compares them on the same targets
using the same ranking rule. Most predictions from the
geometric student avoid counted geometry violations altogether.
The $\rho=7$ student's defects are spread across many more targets,
even though the mean clash counts are close.

Indeed, an average alone misses both parts of this result: how often a
structure is damaged and how badly it is damaged when it fails.
The paired accuracy changes and the violation tail show the remaining
trade-off. Appendices~\ref{app:tails} and~\ref{app:geometry-full} examine how target
size and individual outliers affect these results.

\section{Conclusion and future work}

This work addresses a central problem in evaluating fast structure prediction: high structural-similarity scores can obscure failures in molecular geometry. We combine a geometric noise schedule with multistep consistency training, physical steering, and reference-free candidate selection. Across AlphaFold 3 and Boltz-2, this approach retains much of the teachers' structural accuracy under stereochemically checked evaluation. With the full inference procedure, most predictions from the sixteen-step AF3 student have no counted geometry defects, and fewer predictions are affected than for the corresponding teacher. Its remaining clash burden is concentrated in a relatively small subset of targets. Several accuracy measures also improve over the teacher; our paired comparisons and uncertainty estimates distinguish these gains from differences that remain uncertain.

Blitz provides a practical alternative to the expensive diffusion stage in workflows that repeatedly call a structure predictor, including protein design, virtual screening, and antibody candidate evaluation. Our speedup applies to this stage rather than the full prediction pipeline; the trunk remains unchanged and still represents a substantial computational cost. Future work could combine few-step diffusion with methods that reduce trunk computation or strengthen physical constraints during sampling.

Our results show that substantial acceleration need not come at the expense of structural accuracy or molecular geometry, provided both are considered in training, inference, and evaluation. Average similarity scores alone cannot establish that quality has been preserved. Evaluation should make clear what is preserved, how often predictions fail, and how severe those failures are. Claims of successful acceleration should account for all three.

\section*{Acknowledgments}
We thank the members of the Kumara and Lai labs for their comments and feedback. We also thank Xiowen You, Ankur Verma, and Daniel Cullina for helpful discussions and suggestions, and Charles Pavloski (NCSA NSF-OAC2320345 \& NSF-OAC2005572) for feedback and troubleshooting. We gratefully acknowledge funding support from the Cornell University BRC Epigenomics Core Facility (RRID:SCR\_021287), the Penn State Institute for Computational and Data Sciences (RRID:SCR\_025154), and Penn State AIMI (RRID:SCR\_022867). Soundar R. Kumara would also like to thank the Allen E. Pearce and Allen M. Pearce Professorship. This project was supported by NIH grant R35GM155380 to W.K.M.L. Computational resources were provided through an NSF ACCESS Maximize allocation (CIS260116).

\section*{Use of Generative AI}

We used Claude and ChatGPT as coding assistants for supporting data-processing, evaluation, scoring, and plotting scripts, and for parts of \method{}, including the training loop in Algorithm~\ref{alg:training}. We also used these tools to correct grammar and improve sentence structure in selected passages. They were not used to generate research ideas, propose or refine hypotheses, design the methodology or experiments, or conduct or interpret analyses. The manuscript's content, structure, and arguments are the authors' own. We take full responsibility for the final text, code, and claims.

\subsection*{Ethics Statement}
This work develops computational methods for accelerating biomolecular structure prediction and does not involve human subjects, personal data, or clinical decision-making. AlphaFold 3 experiments were conducted in a non-commercial academic setting using model parameters obtained through the official access process and subject to the applicable terms of use. We do not redistribute AlphaFold 3 model parameters or AF3-derived Blitz checkpoints; users of the released AF3-related code must obtain authorized parameters independently. Boltz2-Blitz parameters and the Blitz training and inference code will be released openly. The methods are intended for scientific research rather than clinical or other consequential decision-making.

\bibliography{references}
\bibliographystyle{blitz_references}
\clearpage
\appendix
\raggedbottom
\renewcommand{\floatpagefraction}{0.7}
\renewcommand{\topfraction}{0.9}
\renewcommand{\bottomfraction}{0.9}
\renewcommand{\textfraction}{0.1}
\makeatletter
\setlength{\@fptop}{0pt}
\setlength{\@fpsep}{16pt}
\setlength{\@fpbot}{0pt plus 1fil}
\makeatother
\section{Training and sampling details}
\label{app:training}

\subsection{Training corpora and optimization}
The AF3 student is initialized from the teacher and distilled on the
publicly available structures from the Protein Data Bank
\citep{berman2000pdb}, re-featurized through the AF3 pipeline. A 640-token crop is taken before featurization.
The Boltz-2 student uses the Boltz data pipeline, sampling RCSB and
OpenFold-processed targets at equal probability with cluster-level
sampling and the Boltz cropper. Evaluation targets are not used for
gradient updates.

The eight-segment training recipe uses an optimization budget of 3,000
steps on 32 H100 GPUs. The fine teacher grid is refined from 200 to 350
edges over training; the student partition remains fixed. The log-normal
weighting has $P_{\mathrm{mean}}=-1.2$, $P_{\mathrm{std}}=1.5$, and
$\sigma_{\mathrm{data}}=16$.
The terminal anchor assigns the final teacher edge to zero noise a proposal
probability of $1/M$, where $M$ is the number of teacher-grid edges,
and rescales the remaining probabilities.

\begin{table}[htbp]
\centering
\caption{Eight-segment Blitz training recipe and AF3 sampling minima.}
\label{tab:training}

\begin{tabular}{ll}
\toprule
Parameter & Setting \\
\midrule
Trainable parameters & Diffusion module only \\
Initialization & Pretrained teacher weights \\
Student segments & 8 \\
Teacher-grid refinement & 200 to 350 edges, log-linear \\
Optimization budget & 3,000 steps \\
Optimizer & Adam/AdamW, no weight decay \\
Peak learning rate & $3\times10^{-5}$ \\
Warmup & 500 steps \\
Adam coefficients & $\beta=(0.9,0.999)$, $\epsilon=10^{-8}$ \\
Effective batch & 128 \\
Diffusion multiplicity & 32, synchronized noise level \\
Recycle updates & 3 \\
Coordinate objective & Weighted alignment and pseudo-Huber \\
Noise weighting & Square-root Karras weighting \\
Terminal anchor & Enabled \\
Dropout & Disabled \\
AF3 relative $\sigma_{\min}$ & 0.01 (4 steps); 0.004 (8 and 16 steps) \\
AF3 absolute $\sigma_{\min}$ & 0.16 / 0.064 / 0.064~\AA{} (4 / 8 / 16 steps) \\
\bottomrule
\end{tabular}
\end{table}

In Algorithm~\ref{alg:training}, $(s,s^-)$ is a sampled adjacent teacher-grid
edge, and $j$ satisfies $b_{j+1}<s\leq b_j$. Edge sampling uses the EDM
proposal above, not uniform sampling over segments. Edge starts coinciding
with an interior student boundary are removed. The teacher destination
is clipped to the lower segment boundary $b=b_{j+1}$.
\textsc{TeacherStep} uses Heun when $s^->0$ and the frozen teacher's
denoised prediction when $s^-=0$. \textsc{Step} applies
Equation~\ref{eq:deployment} with the current student at $S=1$.
When the teacher reaches $b$, its endpoint is used directly. The complete target is stopped-gradient, and neither
branch evaluates a denoiser at zero noise.

For AF3, $\operatorname{Align}_a$ follows the weighted rigid-alignment
procedure of AlphaFold~3 \citep{abramson2024af3}. Our distillation loss
uses a vector pseudo-Huber penalty and square-root Karras weight:
\[
 \operatorname{PH}_{a,\delta}(e)
 =\sqrt{\sum_i a_i\|e_i\|_2^2+\delta^2}-\delta,
 \qquad
 w(s)=\frac{\sqrt{s^2+\sigma_{\mathrm{data}}^2}}{s\sigma_{\mathrm{data}}}.
\]
Here $i$ indexes atoms, $\delta=10^{-4}$, and $\sigma_{\mathrm{data}}=16$.
The reported eight-step AF3 run uses unit weights on resolved atoms,
without chain-type upweighting. The vector penalty is not divided by the
atom count. Losses are averaged over augmented copies and the effective
batch before an Adam update; the copies of each structure share one sampled
edge but have independent augmentation and Gaussian noise.

\subsection{Positive nodes and deployment updates}
\label{app:contracts}
The terminal zero is a projection endpoint, not a positive
log-noise interval.

Both deployed students use deterministic frozen-denoiser exponential
updates. For a positive-noise interval, let $r=b/s$. Holding the
denoiser fixed while integrating in log noise gives
\begin{equation}
 x_b=r^Sx_s+(1-r^S)D_\theta(x_s,s;c).
 \label{eq:deployment}
\end{equation}
We tuned the steering strengths and noise minima on a 2025 PDB validation
set constructed using the same protocol as the Boltz-2 benchmark
\citep{passaro2025boltz2}.
The step scale is $S=1.2$ for AF3 and $S=1.5$ for Boltz-2. At $b=0$,
the limiting update is the denoiser projection. This is a closed-form
update for the frozen-denoiser subproblem, not an exact integration of
the full neural ODE. The controlled numerical experiment instead uses
$S=1$ Euler updates. The matched default and geometric AF3 students
use the same deployed update, while the teachers retain their own
deterministic or stochastic 200-step recipes.

\paragraph{Reference-free candidate ranking.}
For candidate $i$, let $B_i$, $A_i$, and $C_i$ be bond, angle, and clash
violation counts from the model-only checker, and let $i_0$ be the
model's highest-confidence candidate. We use
\[
 G_i=B_i+A_i+C_i,\qquad
 i_g=\mathop{\arg\min}_{i}^{\mathrm{lex}}(G_i,B_i+A_i,i),\qquad
 i^*=\begin{cases}
 i_g,&G_{i_0}-G_{i_g}\geq1,\\
 i_0,&\text{otherwise}.
 \end{cases}
\]
The lexicographic order breaks ties by fewer bond-plus-angle violations,
then lower sample index. If no candidate has fewer total violations than
$i_0$, we retain $i_0$. All counts use predicted coordinates and chemical
identities, without a reference structure.

\section{Datasets and evaluation conventions}
\label{app:datasets}\label{app:evaluation}

\paragraph{Dataset differences.}
We use a reconstruction of the Boltz-2 benchmark
($\sim$2,300 targets) \citep{passaro2025boltz2}.
Boltz-2 is excluded from the Runs~N'~Poses structural comparison because
its training extends beyond that benchmark's cutoff \citep{runsnposes}.
The RecentPDB384 results use the Protenix release
\citep{abramson2024af3,chen2025protenix}; differing model training cutoffs
also preclude treating every comparison as zero-shot.

\paragraph{Scoring choices.}
We use OpenStructure's structural and chemical scores
\citep{biasini2013openstructure,ost211scoring}, with stereochemical
checks enabled for complex lDDT unless marked unchecked
\citep{mariani2013lddt}. PXM is the separate Protenix-native evaluator
\citep{chen2025protenix}.
Table~\ref{tab:conventions} pairs checked and unchecked scores on matched
targets. PXM is evaluated with stereochemical checks disabled.
The broad tables average interface scores over the full evaluated set;
the paired DockQ analysis uses reference-interface-applicable targets
and penalizes missing predicted interfaces.

RF-valid is the percentage of retained predictions passing Boltz's
reference-free physical-validity checks: inter-chain clashes, ligand
bond lengths and angles, internal clashes, stereochemistry, and planarity
\citep{passaro2025boltz2}.

Ligand RMSD and lDDT-PLI average assigned ligands within each target,
then weight targets equally. PB-valid and Success use a fixed primary
ligand under the DeCAF-compatible policy
\citep{scarpellini2026decaf}; Success cannot be reconstructed
from the adjacent aggregate RMSD and PB-valid columns. Primary-ligand
accuracy in Table~\ref{tab:other-samplers} uses PB-evaluable targets.

\section{Benchmark composition}
\label{app:full2269}\label{app:extra}
\subsection{Molecular composition}
\label{app:composition}
Table~\ref{tab:composition} separates molecular compositions on the Boltz-2 benchmark
and RecentPDB384, with matched targets within each comparison.
Protein and protein--ligand complexes dominate the aggregate;
nucleic-acid strata expose differences that it can hide.
\begingroup
\fontsize{10}{12}\selectfont
\setlength{\tabcolsep}{6pt}
\begin{longtable}{@{}lrrrr@{}}
\caption{Stereochecked complex lDDT ($\uparrow$) by molecular composition in protein-containing complexes. Boltz-2 benchmark ($\sim$2,300 targets).}\label{tab:composition}\label{tab:recentpdbcomp}\\
\toprule
Model & P & P+L & P+NA & P+NA+L \\ \midrule\endfirsthead
\multicolumn{5}{c}{\tablename\ \thetable\ (continued): Boltz-2 benchmark ($\sim$2,300 targets)}\\
\toprule
Model & P & P+L & P+NA & P+NA+L \\ \midrule\endhead
\midrule\multicolumn{5}{r}{\textit{Continued on next page}}\\\endfoot
\bottomrule\endlastfoot
AlphaFold 3 teacher & 0.8510 & 0.8841 & 0.7686 & 0.7190 \\
AlphaFold 3 ($\rho{=}7$) & 0.8506 & 0.8835 & 0.7571 & 0.7040 \\
AlphaFold 3-Blitz [4] & 0.8447 & 0.8820 & 0.7454 & 0.6922 \\
AlphaFold 3-Blitz [8] & \textbf{0.8537} & \textbf{0.8864} & 0.7647 & 0.7134 \\
AlphaFold 3-Blitz [16] & \textbf{0.8544} & \textbf{0.8852} & \textbf{0.7696} & 0.7165 \\
Boltz-2 teacher & 0.8432 & 0.8698 & 0.7563 & 0.7066 \\
Boltz-2-Blitz [8] & 0.8405 & \textbf{0.8728} & 0.7305 & 0.6840 \\
Boltz-2-Blitz [16] & \textbf{0.8464} & \textbf{0.8746} & 0.7473 & 0.6989 \\
Boltz-1x teacher & 0.7989 & 0.8344 & 0.6099 & 0.5901 \\
DeCAF default & 0.6580 & 0.7126 & 0.4833 & 0.4259 \\
DeCAF SDE & 0.6429 & 0.7023 & 0.4685 & 0.4138 \\
DeCAF FK & 0.6525 & 0.7051 & 0.4656 & 0.4142 \\
Protenix & 0.8426 & 0.8778 & 0.7806 & 0.7248 \\
Protenix ODE & 0.5259 & 0.5888 & 0.3673 & 0.2887 \\
Protenix Mini & 0.6384 & 0.6905 & 0.5187 & 0.4865 \\
\end{longtable}
\addtocounter{table}{-1}
\renewcommand{\theHtable}{\arabic{table}.panel2}
\begin{longtable}{@{}lrrrr@{}}
\caption{Stereochecked complex lDDT ($\uparrow$) by molecular composition in protein-containing complexes. RecentPDB384.}\\
\toprule
Model & P & P+L & P+NA & P+NA+L \\ \midrule\endfirsthead
\multicolumn{5}{c}{\tablename\ \thetable\ (continued): RecentPDB384}\\
\toprule
Model & P & P+L & P+NA & P+NA+L \\ \midrule\endhead
\midrule\multicolumn{5}{r}{\textit{Continued on next page}}\\\endfoot
\bottomrule\endlastfoot
AlphaFold 3 teacher & 0.8220 & 0.8760 & 0.7786 & 0.7933 \\
AlphaFold 3-Blitz [8] & \textbf{0.8232} & \textbf{0.8776} & \textbf{0.7844} & 0.7927 \\
AlphaFold 3-Blitz [16] & \textbf{0.8232} & \textbf{0.8766} & \textbf{0.7866} & 0.7923 \\
Boltz-1x teacher & 0.7651 & 0.8144 & 0.6560 & 0.6844 \\
DeCAF default & 0.6270 & 0.6934 & 0.5146 & 0.5598 \\
DeCAF FK & 0.6255 & 0.6855 & 0.5109 & 0.5433 \\
DeCAF SDE & 0.6249 & 0.6799 & 0.5001 & 0.5317 \\
Protenix & 0.8073 & 0.8619 & 0.7856 & 0.8060 \\
Protenix ODE & 0.4669 & 0.5590 & 0.3777 & 0.4025 \\
Protenix Mini & 0.5937 & 0.6710 & 0.5117 & 0.5381 \\
\end{longtable}
\par\smallskip\parbox{\linewidth}{P: protein; L: ligand; NA: DNA or RNA, including complexes containing both. Nucleic-acid strata are pooled with equal weight per target. Complexes without protein remain in the overall results.}
\endgroup

\section{Runs N' Poses results}
\label{app:rnp}
Both AF3 students retain structural accuracy;
the sixteen-step student reduces the eight-step geometry errors but
retains more than the teacher.
\begingroup
\fontsize{10}{12}\selectfont
\setlength{\tabcolsep}{3pt}
\begin{longtable}{lrrrrrrrr@{}}
\caption{Runs~N'~Poses: structural accuracy, ligand placement, and geometry.}\label{tab:rnp}\\
\toprule
 & \multicolumn{3}{c}{AF3} & Boltz & \multicolumn{2}{c}{DeCAF} & \multicolumn{2}{c}{Protenix} \\
\cmidrule(lr){2-4}\cmidrule(lr){6-7}\cmidrule(lr){8-9}
Metric & Teacher & \textbf{\shortstack{Blitz\\{[8]}}} & \textbf{\shortstack{Blitz\\{[16]}}} & 1x & Def. & FK & Def. & Mini \\ \midrule\endfirsthead
\multicolumn{9}{c}{\tablename\ \thetable\ (continued)}\\
\toprule
 & \multicolumn{3}{c}{AF3} & Boltz & \multicolumn{2}{c}{DeCAF} & \multicolumn{2}{c}{Protenix} \\
\cmidrule(lr){2-4}\cmidrule(lr){6-7}\cmidrule(lr){8-9}
Metric & Teacher & \textbf{\shortstack{Blitz\\{[8]}}} & \textbf{\shortstack{Blitz\\{[16]}}} & 1x & Def. & FK & Def. & Mini \\ \midrule\endhead
\midrule\multicolumn{9}{r}{\textit{Continued on next page}}\\\endfoot
\bottomrule\endlastfoot
cLDDT $\uparrow$ & .8320 & .8304 & \textbf{.8345} & .8087 & .6914 & .6955 & .8298 & .6451 \\
bbLDDT $\uparrow$ & .8685 & \textbf{.8722} & \textbf{.8712} & .8790 & .8712 & .8640 & .8704 & .8538 \\
TM $\uparrow$ & .8756 & \textbf{.8830} & \textbf{.8777} & .8881 & .8850 & .8826 & .8781 & .8635 \\
QS $\uparrow$ & .5748 & \textbf{.5755} & \textbf{.5753} & .6060 & .6058 & .6055 & .5995 & .5617 \\
DockQ $\uparrow$ & .1486 & \textbf{.1495} & .1484 & .1354 & .1323 & .1303 & .1302 & \textbf{.1385} \\
ICS $\uparrow$ & .1258 & \textbf{.1260} & \textbf{.1265} & .1185 & .1182 & .1176 & .1121 & \textbf{.1150} \\
IPS $\uparrow$ & .1357 & .1312 & .1341 & .1257 & .1249 & .1226 & .1248 & \textbf{.1249} \\
\midrule
lDDT-PLI $\uparrow$ & .6752 & \textbf{.6852} & \textbf{.6783} & .7231 & \textbf{.7608} & \textbf{.7278} & .7634 & .7337 \\
RMSD (\AA) $\downarrow$ & 4.620 & \textbf{4.279} & 4.704 & 3.276 & \textbf{2.656} & \textbf{3.252} & 2.964 & 3.00 \\
Assigned $<2$\AA{} (\%) $\uparrow$ & 62.14 & \textbf{63.17} & \textbf{62.48} & 70.98 & \textbf{76.44} & 70.81 & 74.49 & 72.05 \\
PB-valid (\%) $\uparrow$ & 72.37 & \textbf{74.47} & \textbf{72.60} & 95.32 & 54.33 & 91.33 & 72.37 & 67.45 \\
95\% CI & \shortstack{[68.1,\\76.6]} & \shortstack{[70.3,\\78.5]} & \shortstack{[68.1,\\76.8]} & \shortstack{[93.2,\\97.2]} & \shortstack{[49.6,\\59.0]} & \shortstack{[88.8,\\93.9]} & \shortstack{[68.1,\\76.6]} & \shortstack{[63.0,\\71.7]} \\
Primary $<2$\AA{} (\%) $\uparrow$ & 83.84 & 83.37 & 83.37 & 78.92 & \textbf{79.86} & \textbf{80.33} & 81.97 & 77.05 \\
95\% CI & \shortstack{[80.3,\\87.1]} & \shortstack{[79.9,\\86.9]} & \shortstack{[79.9,\\86.9]} & \shortstack{[74.9,\\82.7]} & \shortstack{[76.1,\\83.6]} & \shortstack{[76.6,\\84.1]} & \shortstack{[78.2,\\85.5]} & \shortstack{[73.1,\\81.0]} \\
Success (\%) $\uparrow$ & 61.36 & \textbf{64.17} & \textbf{62.06} & 75.88 & 47.54 & 73.30 & 60.42 & 54.80 \\
95\% CI & \shortstack{[56.7,\\66.0]} & \shortstack{[59.5,\\68.6]} & \shortstack{[57.4,\\66.7]} & \shortstack{[71.9,\\79.9]} & \shortstack{[42.9,\\52.2]} & \shortstack{[69.1,\\77.5]} & \shortstack{[56.0,\\65.1]} & \shortstack{[50.1,\\59.5]} \\
RF-valid (\%) $\uparrow$ & 13.37 & 11.94 & 13.07 & 47.47 & 4.25 & 26.44 & 11.09 & 5.45 \\
\midrule
Bonds $\downarrow$ & 0.01 & 4.04 & 0.68 & 9.96 & 317.15 & 326.62 & 0.13 & 452.59 \\
Angles $\downarrow$ & 0.01 & 0.65 & 0.11 & 3.90 & 142.02 & 155.58 & 0.26 & 149.52 \\
Clashes $\downarrow$ & 0.36 & 19.84 & 2.74 & 54.03 & 415.79 & 336.86 & 4.30 & 308.85 \\
Violations $\downarrow$ & 0.38 & 24.53 & 3.54 & 67.89 & 874.96 & 819.05 & 4.70 & 910.97 \\
OST-clean (\%) $\uparrow$ & 91.73 & 85.55 & 88.39 & 3.09 & 0.00 & 0.00 & 50.38 & 0.00 \\
\end{longtable}
\par\smallskip\parbox{\linewidth}{PB-valid, primary accuracy, and Success all use the same 427 targets (17\% of the eligible set). Ligand-assignment and scoring failures restrict this subset, so these rates are not full-benchmark estimates. Brackets give 95\% bootstrap intervals. DeCAF FK uses four particles and twelve steps. }
\endgroup

\section{RecentPDB384 results}
\label{app:recentpdb}
The sixteen-step student reduces the eight-step geometry errors while
retaining similar structural accuracy. We exclude Boltz-2 and its students
because this benchmark predates its June 2023 training cutoff
\citep{chen2025protenix,passaro2025boltz2}.
\begingroup
\fontsize{10}{12}\selectfont
\setlength{\tabcolsep}{3pt}
\begin{longtable}{lrrrrrrrrrr@{}}
\caption{RecentPDB384: structural accuracy, ligand placement, and geometry.}\label{tab:recentpdb}\\
\toprule
 & \multicolumn{3}{c}{AF3} & Boltz & \multicolumn{3}{c}{DeCAF} & \multicolumn{3}{c}{Protenix} \\
\cmidrule(lr){2-4}\cmidrule(lr){6-8}\cmidrule(lr){9-11}
Metric & Teacher & \textbf{\shortstack{Blitz\\{[8]}}} & \textbf{\shortstack{Blitz\\{[16]}}} & 1x & Def. & FK & SDE & Def. & ODE & Mini \\ \midrule\endfirsthead
\multicolumn{11}{c}{\tablename\ \thetable\ (continued)}\\
\toprule
 & \multicolumn{3}{c}{AF3} & Boltz & \multicolumn{3}{c}{DeCAF} & \multicolumn{3}{c}{Protenix} \\
\cmidrule(lr){2-4}\cmidrule(lr){6-8}\cmidrule(lr){9-11}
Metric & Teacher & \textbf{\shortstack{Blitz\\{[8]}}} & \textbf{\shortstack{Blitz\\{[16]}}} & 1x & Def. & FK & SDE & Def. & ODE & Mini \\ \midrule\endhead
\midrule\multicolumn{11}{r}{\textit{Continued on next page}}\\\endfoot
\bottomrule\endlastfoot
cLDDT $\uparrow$ & .8286 & .8273 & .8286 & .7296 & .6051 & .5993 & .5954 & .8231 & .4680 & .5979 \\
bbLDDT $\uparrow$ & .8948 & \textbf{.8961} & \textbf{.8958} & .8673 & .8637 & .8573 & .8573 & .8866 & .8822 & .8683 \\
TM $\uparrow$ & .8572 & \textbf{.8590} & .8567 & .8287 & .8240 & .8191 & .8226 & .8500 & .8486 & .8277 \\
QS $\uparrow$ & .7482 & \textbf{.7502} & .7466 & .6986 & .6922 & .6959 & .6891 & .7257 & .7167 & .6901 \\
DockQ $\uparrow$ & .3428 & \textbf{.3431} & .3395 & .3036 & .2969 & .2936 & .2957 & .3258 & \textbf{.3275} & .2986 \\
ICS $\uparrow$ & .3704 & \textbf{.3715} & .3692 & .3207 & .3163 & .3182 & .3137 & .3430 & .3371 & .3220 \\
IPS $\uparrow$ & .4138 & \textbf{.4145} & \textbf{.4147} & .3763 & .3760 & .3724 & .3723 & .3962 & .3930 & .3817 \\
\midrule
lDDT-PLI $\uparrow$ & .5363 & .5342 & \textbf{.5364} & .4948 & \textbf{.5039} & \textbf{.4977} & .4898 & .5257 & .4844 & .5055 \\
RMSD (\AA) $\downarrow$ & 9.147 & \textbf{8.480} & \textbf{8.473} & 9.159 & \textbf{7.982} & \textbf{8.405} & \textbf{8.751} & 8.086 & \textbf{7.967} & \textbf{8.08} \\
Assigned $<2$\AA{} (\%) $\uparrow$ & 49.28 & 47.55 & 47.34 & 44.72 & 44.71 & 42.84 & 42.16 & 48.10 & 42.79 & 42.90 \\
PB-valid (\%) $\uparrow$ & 51.14 & 47.73 & 45.45 & 77.27 & 22.73 & 71.59 & 19.32 & 55.68 & 28.41 & 43.18 \\
Primary $<2$\AA{} (\%) $\uparrow$ & 61.36 & 59.09 & 60.23 & 57.95 & 56.82 & 56.82 & 56.82 & 63.64 & 60.23 & 57.95 \\
Success (\%) $\uparrow$ & 30.68 & \textbf{32.95} & 29.55 & 46.59 & 17.05 & 45.45 & 14.77 & 39.77 & 28.41 & 37.50 \\
RF-valid (\%) $\uparrow$ & 40.36 & 39.32 & 37.50 & 92.45 & 25.78 & 61.98 & 25.78 & 34.11 & 28.91 & 32.55 \\
\midrule
Bonds $\downarrow$ & 0.52 & 2.29 & 0.80 & 15.76 & 254.24 & 263.73 & 256.59 & 0.62 & 729.90 & 382.14 \\
Angles $\downarrow$ & 0.47 & 1.10 & 0.54 & 8.47 & 119.82 & 132.35 & 128.16 & 0.67 & 326.12 & 136.47 \\
Clashes $\downarrow$ & 1.61 & 4.35 & 2.80 & 19.96 & 254.65 & 192.72 & 253.66 & 5.27 & 1001.85 & 195.19 \\
Violations $\downarrow$ & 2.61 & 7.74 & 4.14 & 44.18 & 628.71 & 588.81 & 638.41 & 6.55 & 2057.87 & 713.80 \\
OST-clean (\%) $\uparrow$ & 68.53 & 67.20 & \textbf{70.93} & 2.67 & 0.00 & 0.00 & 0.00 & 38.93 & 0.00 & 0.00 \\
PXM $\uparrow$ & .8260 & \textbf{.8286} & \textbf{.8278} & .7884 & .7872 & .7804 & .7808 & .8206 & .8192 & .7942 \\
\end{longtable}

\endgroup

\raggedbottom
\section{Size and tail sensitivity in the matched AF3 comparison}
\label{app:tails}
\begin{figure}[!htbp]
\centering\includegraphics[width=.8\linewidth]{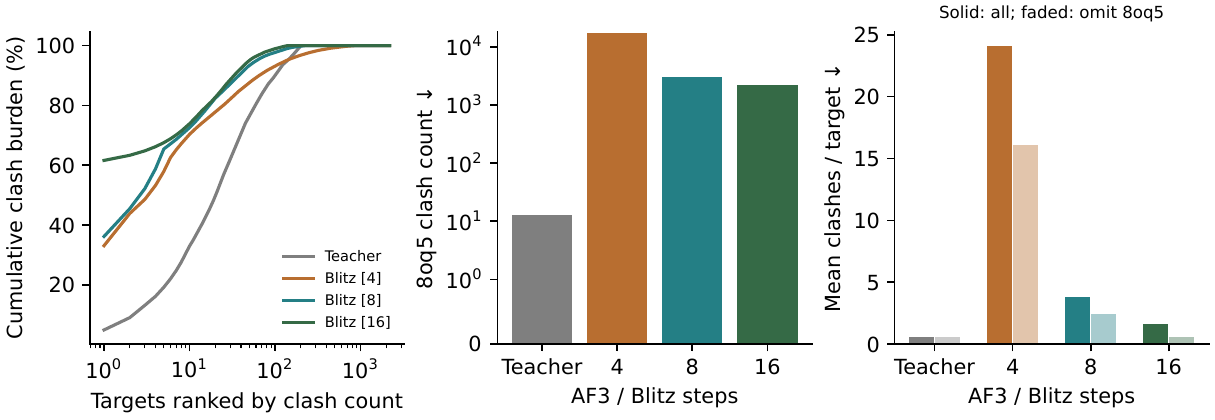}
\caption{\textbf{A small subset carries much of the clash burden.}
Left: cumulative clashes ranked by target. Middle: counts for 8oq5.
Right: means with all targets and after omitting 8oq5 as an influence
diagnostic. Every primary result retains this target.}
\label{fig:geometry-influence}
\end{figure}

A small subset of targets contributes much of the remaining clash burden.
For the sixteen-step AF3 student, 8oq5 alone accounts for 61.6\% of
counted clashes on the matched comparison (Figure~\ref{fig:geometry-influence}).
We retain this target in every primary result. The influence analysis
shows why mean counts should be read alongside defect incidence and
the full distribution, rather than used as a description of a typical prediction.

\section{Geometry and candidate selection}
\label{app:geometry-full}
We compare the canonical stochastic teacher with the four-, eight-, and
sixteen-step students on matched Boltz-2 benchmark targets. The eight-step schedule
comparison is shown in Figure~\ref{fig:schedule-violations}.

\subsection{How often a defect occurs and how severe it becomes}
\begin{figure}[!htbp]
\centering\includegraphics[width=.86\linewidth]{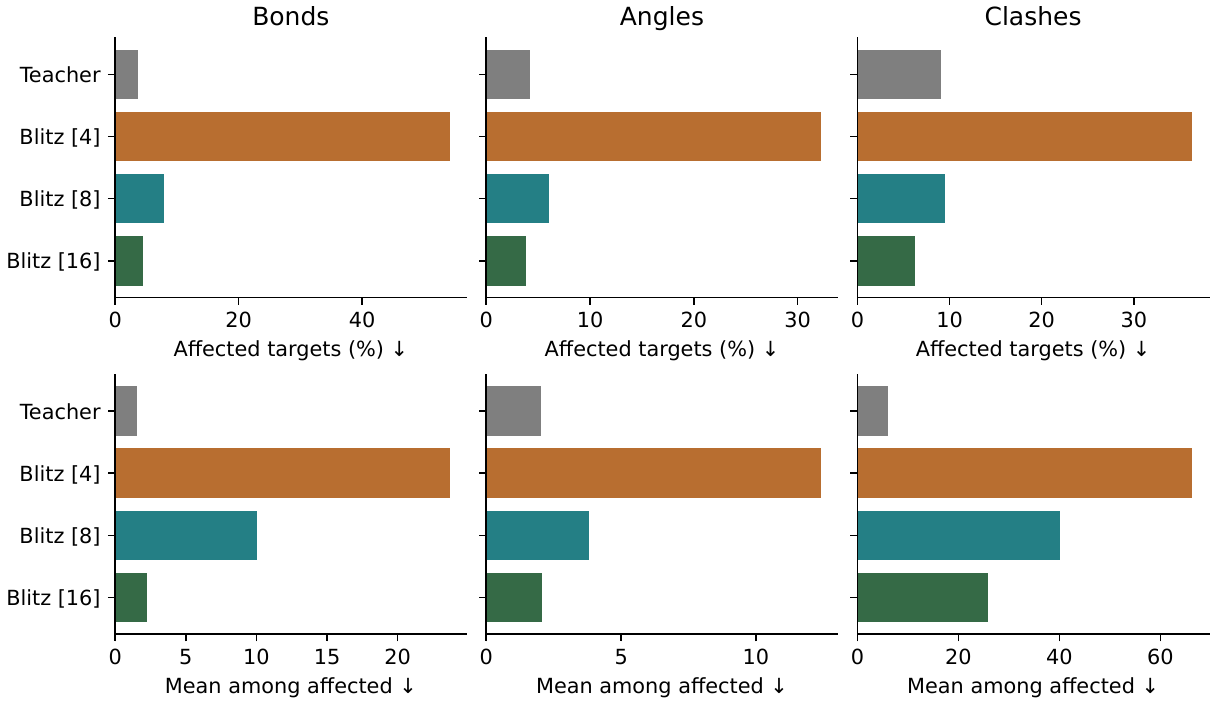}
\caption{\textbf{Severity and incidence answer different questions.}
Top: targets with a counted defect. Bottom: mean count among affected targets.}
\label{fig:geometry-incidence}
\end{figure}

A small number of heavily damaged predictions can dominate a mean count.
Figure~\ref{fig:geometry-incidence} separates the fraction of affected targets
from the mean number of violations among those targets.
The sixteen-step student has fewer defect-affected predictions than the
teacher in this comparison, but its remaining defects are more concentrated.
A low incidence therefore does not imply low severity when a failure occurs.
These checks cover polymer bonds, angles, and clashes; passing them does
not establish full chemical validity or replace ligand PoseBusters checks.

\subsection{Distribution tails and target-size sensitivity}
\begin{figure}[!htbp]
\centering\includegraphics[width=.86\linewidth]{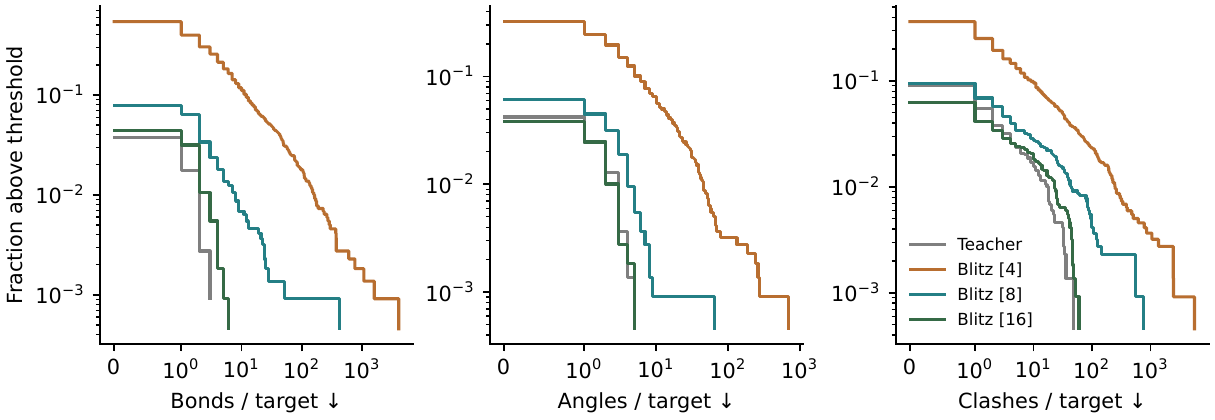}
\caption{\textbf{The full violation distribution.}
Curves show the fraction of matched targets above each count. The value
at zero is incidence; no extreme target is removed.}
\label{fig:geometry-distribution}
\end{figure}
\begin{figure}[!htbp]
\centering\includegraphics[width=.86\linewidth]{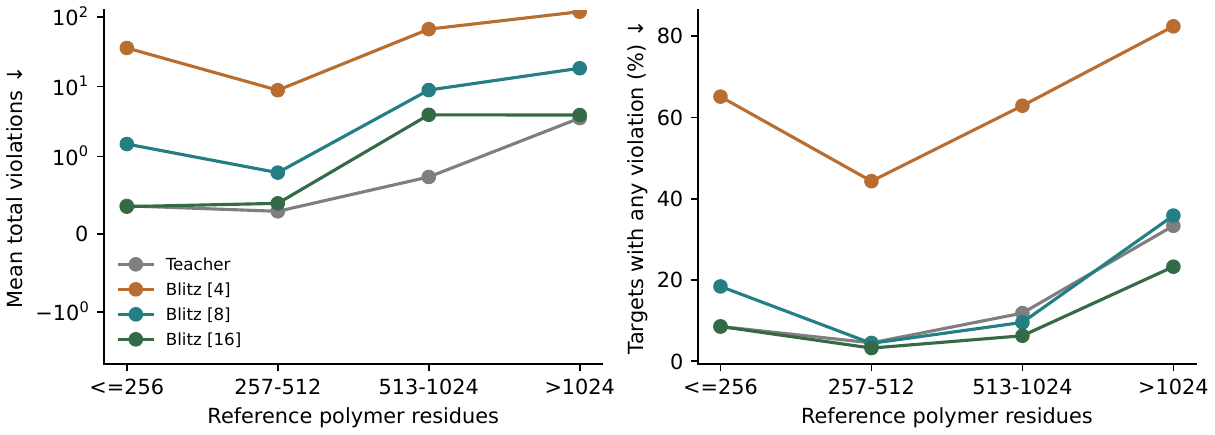}
\caption{\textbf{Geometry across target sizes.}
Mean total count and defect incidence within predefined reference-residue
bins, using identical targets for all four methods.}
\label{fig:geometry-size}
\end{figure}
Figures~\ref{fig:geometry-distribution} and~\ref{fig:geometry-size} show
how much of the mean comes from the tail and whether the pattern persists
across target sizes. Each target has equal weight within its size bin.

\subsection{Effect of candidate selection}
\label{app:native-selection}
Table~\ref{tab:native-selection} uses each model's native confidence ranking.
Structural accuracy remains competitive, and some ligand and interface
scores improve relative to our reranker, but geometry violations are more
frequent. On identical targets, reranking reduces the affected fraction
from 13.9\% to 6.9\% for AF3-Blitz [16] and from 42.6\% to 26.9\%
for Boltz-2-Blitz [16]. Checked lDDT also improves in these paired
comparisons. We therefore retain the reranker: the lower defect incidence
than the teacher reported for the sixteen-step students depends on this
selection rule.

\begingroup
\fontsize{10}{12}\selectfont
\setlength{\tabcolsep}{2pt}
\setlength{\LTleft}{\fill}
\setlength{\LTright}{\fill}
\setlength{\LTpre}{8pt}
\setlength{\LTpost}{8pt}
\begin{longtable}{@{}lrrrrrrrr@{}}
\caption{Native-confidence selection on the Boltz-2 benchmark.}\label{tab:native-selection}\\
\toprule
& \multicolumn{5}{c}{AlphaFold3} & \multicolumn{3}{c}{Boltz-2} \\
\cmidrule(lr){2-6}\cmidrule(lr){7-9}
Metric & Teacher & $\rho{=}7$ & \shortstack{Blitz\\{[4]}} & \shortstack{Blitz\\{[8]}} & \shortstack{Blitz\\{[16]}} & Teacher & \shortstack{Blitz\\{[8]}} & \shortstack{Blitz\\{[16]}} \\
\midrule
\endfirsthead
\multicolumn{9}{c}{\tablename~\thetable{} (continued): Native-confidence selection}\\
\toprule
& \multicolumn{5}{c}{AlphaFold3} & \multicolumn{3}{c}{Boltz-2} \\
\cmidrule(lr){2-6}\cmidrule(lr){7-9}
Metric & Teacher & $\rho{=}7$ & \shortstack{Blitz\\{[4]}} & \shortstack{Blitz\\{[8]}} & \shortstack{Blitz\\{[16]}} & Teacher & \shortstack{Blitz\\{[8]}} & \shortstack{Blitz\\{[16]}} \\
\midrule
\endhead
\midrule
\multicolumn{9}{r}{\emph{Continued on next page}}\\
\endfoot
\bottomrule
\endlastfoot
NFE & 200 & 8 & 4 & 8 & 16 & 600 & 8 & 16 \\
\midrule
cLDDT $\uparrow$ & 0.8615 & 0.8550 & 0.8512 & 0.8615 & \textbf{0.8624} & 0.8503 & 0.8476 & \textbf{0.8525} \\
bbLDDT $\uparrow$ & 0.9136 & \textbf{0.9147} & 0.9128 & \textbf{0.9144} & \textbf{0.9150} & 0.9047 & \textbf{0.9061} & \textbf{0.9074} \\
TM $\uparrow$ & 0.8898 & \textbf{0.8902} & 0.8898 & \textbf{0.8901} & \textbf{0.8909} & 0.8758 & \textbf{0.8789} & \textbf{0.8792} \\
DockQ $\uparrow$ & 0.4198 & 0.4163 & \textbf{0.4220} & \textbf{0.4216} & 0.4168 & 0.3927 & \textbf{0.3929} & \textbf{0.3948} \\
QS $\uparrow$ & 0.7784 & 0.7763 & \textbf{0.7793} & 0.7767 & 0.7782 & 0.7378 & \textbf{0.7381} & \textbf{0.7381} \\
ICS $\uparrow$ & 0.4263 & 0.4239 & \textbf{0.4274} & 0.4258 & 0.4245 & 0.3883 & 0.3880 & 0.3878 \\
IPS $\uparrow$ & 0.4560 & \textbf{0.4567} & \textbf{0.4578} & \textbf{0.4572} & \textbf{0.4581} & 0.4223 & \textbf{0.4256} & \textbf{0.4252} \\
\midrule
lDDT-PLI $\uparrow$ & 0.6437 & 0.6435 & 0.6386 & \textbf{0.6448} & \textbf{0.6471} & 0.5596 & 0.5572 & \textbf{0.5598} \\
RMSD (\AA{}) $\downarrow$ & 6.63 & \textbf{6.51} & \textbf{6.43} & \textbf{6.43} & \textbf{6.38} & 9.00 & \textbf{8.59} & \textbf{8.67} \\
Assigned $<2$ \AA{} (\%) $\uparrow$ & 61.3 & \textbf{61.4} & 60.8 & \textbf{61.6} & \textbf{61.8} & 52.2 & 52.0 & \textbf{52.7} \\
PB-valid (\%) $\uparrow$ & 45.1 & 38.4 & 35.3 & 42.9 & \textbf{45.7} & 60.9 & 53.1 & 55.5 \\
Primary $<2$ \AA{} (\%) $\uparrow$ & 70.8 & \textbf{71.4} & \textbf{71.2} & \textbf{71.8} & \textbf{71.9} & 61.5 & 61.2 & \textbf{61.9} \\
Success (\%) $\uparrow$ & 34.7 & 32.5 & 31.2 & \textbf{35.7} & \textbf{36.2} & 36.5 & 31.1 & 34.6 \\
RF-valid (\%) $\uparrow$ & 41.5 & 37.4 & 36.7 & 39.5 & \textbf{41.9} & 98.5 & 93.6 & 93.7 \\
\midrule
Bonds $\downarrow$ & 0.06 & 6.46 & 23.36 & 2.21 & 0.21 & 0.82 & 7.45 & 2.58 \\
Angles $\downarrow$ & 0.09 & 3.04 & 8.58 & 0.67 & 0.17 & 0.61 & 4.81 & 1.86 \\
Clashes $\downarrow$ & 0.54 & 7.87 & 49.80 & 7.71 & 2.89 & 2.52 & 9.79 & 4.64 \\
Any violation (\%) $\downarrow$ & 9.8 & 80.2 & 83.9 & 27.9 & 13.8 & 30.2 & 62.4 & 42.5 \\
\end{longtable}
\noindent Geometry counts are per target. Comparisons use matched evaluable targets within each family; ligand subsets can differ from Table~\ref{tab:af3}. \par
\endgroup

\section{Uncertainty estimates}
\label{app:uncertainty}
Table~\ref{tab:uncertainty} reports pointwise 95\% intervals from 10,000
PDB-cluster bootstrap draws, measuring target-sampling uncertainty rather
than variation across training runs.

\begingroup
\fontsize{10}{12}\selectfont
\setlength{\tabcolsep}{3pt}
\setlength{\LTleft}{\fill}
\setlength{\LTright}{\fill}
\setlength{\LTpre}{8pt}
\setlength{\LTpost}{8pt}
\begin{longtable}{@{}lrrrrr@{}}
\caption{Uncertainty in mean scores and paired comparisons. A. Mean uncertainty: AF3.}\label{tab:uncertainty}\\
\toprule
Metric & Teacher & $\rho=7$ & Blitz [4] & Blitz [8] & Blitz [16] \\
\midrule
\endfirsthead
\multicolumn{6}{c}{\tablename~\thetable{} (continued): A. Mean uncertainty: AF3}\\
\toprule
Metric & Teacher & $\rho=7$ & Blitz [4] & Blitz [8] & Blitz [16] \\
\midrule
\endhead
\midrule
\multicolumn{6}{r}{\emph{Continued on next page}}\\
\endfoot
\bottomrule
\endlastfoot
cLDDT $\uparrow$ & [0.857, 0.866] & [0.855, 0.865] & [0.850, 0.861] & [0.859, 0.868] & [0.859, 0.868] \\
bbLDDT $\uparrow$ & [0.909, 0.918] & [0.911, 0.919] & [0.908, 0.917] & [0.910, 0.919] & [0.911, 0.919] \\
TM $\uparrow$ & [0.882, 0.897] & [0.881, 0.896] & [0.881, 0.896] & [0.881, 0.896] & [0.882, 0.897] \\
DockQ $\uparrow$ & [0.403, 0.437] & [0.397, 0.430] & [0.402, 0.435] & [0.404, 0.438] & [0.400, 0.433] \\
QS $\uparrow$ & [0.765, 0.791] & [0.763, 0.788] & [0.767, 0.794] & [0.763, 0.790] & [0.765, 0.791] \\
ICS $\uparrow$ & [0.410, 0.443] & [0.405, 0.438] & [0.412, 0.446] & [0.410, 0.443] & [0.408, 0.441] \\
IPS $\uparrow$ & [0.439, 0.472] & [0.438, 0.470] & [0.441, 0.474] & [0.441, 0.474] & [0.440, 0.473] \\
lDDT-PLI $\uparrow$ & [0.625, 0.662] & [0.625, 0.662] & [0.619, 0.656] & [0.624, 0.661] & [0.625, 0.662] \\
RMSD (\AA) $\downarrow$ & [6.06, 7.21] & [5.97, 7.10] & [5.90, 7.00] & [5.90, 7.00] & [5.96, 7.08] \\
Assigned $<2$\AA{} $\uparrow$ & [58.9, 63.7] & [58.9, 63.8] & [58.6, 63.4] & [58.9, 63.7] & [58.9, 63.8] \\
PB-valid $\uparrow$ & [42.0, 48.0] & [32.3, 38.1] & [31.8, 37.5] & [40.0, 46.1] & [42.5, 48.7] \\
Primary $<2$\AA{} $\uparrow$ & [65.2, 70.6] & [64.6, 70.0] & [64.8, 70.2] & [65.2, 70.7] & [65.6, 71.0] \\
Success $\uparrow$ & [31.8, 37.6] & [26.7, 32.4] & [27.5, 33.2] & [32.8, 38.7] & [32.9, 38.8] \\
RF-valid $\uparrow$ & [39.4, 43.6] & [34.4, 38.3] & [34.9, 39.0] & [38.5, 42.5] & [40.3, 44.3] \\
Bonds $\downarrow$ & [0.05, 0.07] & [2.39, 3.39] & [7.75, 19.76] & [0.29, 1.57] & [0.08, 0.13] \\
Angles $\downarrow$ & [0.07, 0.11] & [0.89, 1.16] & [2.53, 6.30] & [0.15, 0.34] & [0.06, 0.10] \\
Clashes $\downarrow$ & [0.41, 0.69] & [1.97, 7.55] & [11.02, 44.46] & [1.61, 7.27] & [0.49, 3.69] \\
Any violation $\downarrow$ & [8.6, 11.1] & [48.5, 52.7] & [55.4, 59.5] & [10.0, 12.7] & [5.8, 7.9] \\
\end{longtable}
\addtocounter{table}{-1}
\renewcommand{\theHtable}{\arabic{table}.panel2}
\begin{longtable}{@{}lrrr@{}}
\caption{Uncertainty in mean scores and paired comparisons. B. Mean uncertainty: Boltz-2.}\\
\toprule
Metric & Teacher & Blitz [8] & Blitz [16] \\
\midrule
\endfirsthead
\multicolumn{4}{c}{\tablename~\thetable{} (continued): B. Mean uncertainty: Boltz-2}\\
\toprule
Metric & Teacher & Blitz [8] & Blitz [16] \\
\midrule
\endhead
\midrule
\multicolumn{4}{r}{\emph{Continued on next page}}\\
\endfoot
\bottomrule
\endlastfoot
cLDDT $\uparrow$ & [0.845, 0.855] & [0.844, 0.854] & [0.848, 0.858] \\
bbLDDT $\uparrow$ & [0.900, 0.909] & [0.901, 0.910] & [0.902, 0.911] \\
TM $\uparrow$ & [0.867, 0.883] & [0.869, 0.885] & [0.870, 0.885] \\
DockQ $\uparrow$ & [0.376, 0.409] & [0.373, 0.406] & [0.377, 0.410] \\
QS $\uparrow$ & [0.724, 0.751] & [0.721, 0.749] & [0.722, 0.749] \\
ICS $\uparrow$ & [0.372, 0.404] & [0.369, 0.401] & [0.370, 0.402] \\
IPS $\uparrow$ & [0.406, 0.438] & [0.406, 0.438] & [0.407, 0.438] \\
lDDT-PLI $\uparrow$ & [0.541, 0.578] & [0.536, 0.573] & [0.538, 0.576] \\
RMSD (\AA) $\downarrow$ & [8.32, 9.70] & [8.11, 9.39] & [8.14, 9.41] \\
Assigned $<2$\AA{} $\uparrow$ & [49.7, 54.6] & [49.2, 54.1] & [49.7, 54.6] \\
PB-valid $\uparrow$ & [58.0, 63.8] & [49.0, 55.1] & [52.9, 59.0] \\
Primary $<2$\AA{} $\uparrow$ & [58.0, 63.8] & [57.3, 63.2] & [57.8, 63.6] \\
Success $\uparrow$ & [33.6, 39.5] & [27.8, 33.4] & [31.1, 36.8] \\
RF-valid $\uparrow$ & [98.0, 99.0] & [92.6, 94.6] & [93.2, 95.1] \\
Bonds $\downarrow$ & [0.74, 1.11] & [3.28, 7.02] & [1.18, 2.67] \\
Angles $\downarrow$ & [0.54, 0.97] & [1.96, 4.61] & [0.77, 1.84] \\
Clashes $\downarrow$ & [2.03, 3.77] & [4.58, 8.48] & [2.58, 4.84] \\
Any violation $\downarrow$ & [28.4, 32.2] & [39.8, 44.0] & [24.9, 28.6] \\
\end{longtable}
\noindent Intervals use 10,000 PDB-cluster bootstrap draws. Rates are percentages; geometry counts are per target.\par
\addtocounter{table}{-1}
\renewcommand{\theHtable}{\arabic{table}.panel3}
\begin{longtable}{@{}lrrrrrrr@{}}
\caption{Uncertainty in mean scores and paired comparisons. C. Mean uncertainty: other methods.}\\
\toprule
Metric & Boltz-1x & DeCAF & \shortstack{DeCAF\\SDE} & \shortstack{DeCAF\\FK} & Protenix & \shortstack{Protenix\\ODE} & \shortstack{Protenix\\Mini} \\
\midrule
\endfirsthead
\multicolumn{8}{c}{\tablename~\thetable{} (continued): C. Mean uncertainty: other methods}\\
\toprule
Metric & Boltz-1x & DeCAF & \shortstack{DeCAF\\SDE} & \shortstack{DeCAF\\FK} & Protenix & \shortstack{Protenix\\ODE} & \shortstack{Protenix\\Mini} \\
\midrule
\endhead
\midrule
\multicolumn{8}{r}{\emph{Continued on next page}}\\
\endfoot
\bottomrule
\endlastfoot
cLDDT $\uparrow$ & \shortstack{[0.799,\\0.811]} & \shortstack{[0.666,\\0.682]} & \shortstack{[0.653,\\0.668]} & \shortstack{[0.658,\\0.673]} & \shortstack{[0.848,\\0.859]} & \shortstack{[0.538,\\0.554]} & \shortstack{[0.651,\\0.663]} \\
bbLDDT $\uparrow$ & \shortstack{[0.888,\\0.897]} & \shortstack{[0.882,\\0.891]} & \shortstack{[0.874,\\0.883]} & \shortstack{[0.874,\\0.884]} & \shortstack{[0.902,\\0.912]} & \shortstack{[0.899,\\0.909]} & \shortstack{[0.889,\\0.899]} \\
TM $\uparrow$ & \shortstack{[0.863,\\0.878]} & \shortstack{[0.854,\\0.870]} & \shortstack{[0.850,\\0.866]} & \shortstack{[0.849,\\0.866]} & \shortstack{[0.873,\\0.888]} & \shortstack{[0.870,\\0.886]} & \shortstack{[0.860,\\0.876]} \\
DockQ $\uparrow$ & \shortstack{[0.372,\\0.403]} & \shortstack{[0.355,\\0.385]} & \shortstack{[0.348,\\0.377]} & \shortstack{[0.349,\\0.379]} & \shortstack{[0.391,\\0.424]} & \shortstack{[0.387,\\0.419]} & \shortstack{[0.353,\\0.384]} \\
QS $\uparrow$ & \shortstack{[0.719,\\0.746]} & \shortstack{[0.710,\\0.738]} & \shortstack{[0.705,\\0.733]} & \shortstack{[0.712,\\0.739]} & \shortstack{[0.743,\\0.769]} & \shortstack{[0.735,\\0.762]} & \shortstack{[0.704,\\0.732]} \\
ICS $\uparrow$ & \shortstack{[0.369,\\0.400]} & \shortstack{[0.359,\\0.389]} & \shortstack{[0.355,\\0.385]} & \shortstack{[0.360,\\0.391]} & \shortstack{[0.391,\\0.424]} & \shortstack{[0.384,\\0.416]} & \shortstack{[0.353,\\0.384]} \\
IPS $\uparrow$ & \shortstack{[0.407,\\0.438]} & \shortstack{[0.402,\\0.433]} & \shortstack{[0.399,\\0.429]} & \shortstack{[0.402,\\0.433]} & \shortstack{[0.424,\\0.455]} & \shortstack{[0.424,\\0.456]} & \shortstack{[0.404,\\0.436]} \\
lDDT-PLI $\uparrow$ & \shortstack{[0.575,\\0.612]} & \shortstack{[0.569,\\0.606]} & \shortstack{[0.563,\\0.599]} & \shortstack{[0.569,\\0.606]} & \shortstack{[0.606,\\0.643]} & \shortstack{[0.558,\\0.598]} & \shortstack{[0.559,\\0.597]} \\
RMSD (\AA) $\downarrow$ & \shortstack{[6.47,\\7.56]} & \shortstack{[6.44,\\7.53]} & \shortstack{[6.58,\\7.66]} & \shortstack{[6.50,\\7.58]} & \shortstack{[6.44,\\7.56]} & \shortstack{[6.19,\\7.19]} & \shortstack{[6.88,\\8.02]} \\
Assigned $<2$\AA{} $\uparrow$ & \shortstack{[54.9,\\59.8]} & \shortstack{[53.5,\\58.5]} & \shortstack{[53.7,\\58.5]} & \shortstack{[54.1,\\59.0]} & \shortstack{[57.0,\\61.9]} & \shortstack{[52.4,\\57.4]} & \shortstack{[51.9,\\56.8]} \\
PB-valid $\uparrow$ & \shortstack{[55.2,\\61.2]} & \shortstack{[22.1,\\27.3]} & \shortstack{[24.1,\\29.4]} & \shortstack{[51.0,\\56.9]} & \shortstack{[42.8,\\48.8]} & \shortstack{[24.0,\\29.4]} & \shortstack{[25.5,\\31.0]} \\
Primary $<2$\AA{} $\uparrow$ & \shortstack{[63.3,\\69.2]} & \shortstack{[62.8,\\68.6]} & \shortstack{[62.9,\\68.7]} & \shortstack{[62.6,\\68.5]} & \shortstack{[66.5,\\72.2]} & \shortstack{[62.8,\\68.6]} & \shortstack{[60.9,\\66.6]} \\
Success $\uparrow$ & \shortstack{[36.4,\\42.3]} & \shortstack{[18.6,\\23.6]} & \shortstack{[19.2,\\24.3]} & \shortstack{[35.4,\\41.3]} & \shortstack{[31.2,\\36.9]} & \shortstack{[22.4,\\27.6]} & \shortstack{[21.5,\\26.6]} \\
RF-valid $\uparrow$ & \shortstack{[98.6,\\99.4]} & \shortstack{[24.2,\\27.7]} & \shortstack{[25.7,\\29.4]} & \shortstack{[64.9,\\68.8]} & \shortstack{[30.8,\\34.7]} & \shortstack{[29.0,\\32.8]} & \shortstack{[29.5,\\33.4]} \\
Bonds $\downarrow$ & \shortstack{[9.47,\\14.45]} & \shortstack{[252.86,\\284.68]} & \shortstack{[256.20,\\287.62]} & \shortstack{[263.42,\\297.15]} & \shortstack{[0.16,\\0.21]} & \shortstack{[637.36,\\696.26]} & \shortstack{[357.66,\\390.61]} \\
Angles $\downarrow$ & \shortstack{[4.10,\\7.30]} & \shortstack{[115.68,\\132.17]} & \shortstack{[123.16,\\140.39]} & \shortstack{[129.07,\\147.97]} & \shortstack{[0.50,\\0.67]} & \shortstack{[269.23,\\299.08]} & \shortstack{[123.68,\\134.82]} \\
Clashes $\downarrow$ & \shortstack{[16.22,\\21.68]} & \shortstack{[258.64,\\319.85]} & \shortstack{[250.01,\\304.12]} & \shortstack{[206.79,\\251.70]} & \shortstack{[2.77,\\5.04]} & \shortstack{[544.28,\\742.67]} & \shortstack{[164.69,\\197.33]} \\
Any violation $\downarrow$ & \shortstack{[94.1,\\95.9]} & \shortstack{[96.3,\\97.7]} & \shortstack{[96.8,\\98.1]} & \shortstack{[97.0,\\98.3]} & \shortstack{[47.0,\\51.2]} & \shortstack{[100.0,\\100.0]} & \shortstack{[99.9,\\100.0]} \\
\end{longtable}
\noindent Intervals use 10,000 PDB-cluster bootstrap draws. Rates are percentages; geometry counts are per target.\par
\addtocounter{table}{-1}
\renewcommand{\theHtable}{\arabic{table}.panel4}
\begin{longtable}{@{}lrrrrr@{}}
\caption{Uncertainty in mean scores and paired comparisons. D. Student minus teacher.}\\
\toprule
Metric & AF3 [4] & AF3 [8] & AF3 [16] & Boltz-2 [8] & Boltz-2 [16] \\
\midrule
\endfirsthead
\multicolumn{6}{c}{\tablename~\thetable{} (continued): D. Student minus teacher}\\
\toprule
Metric & AF3 [4] & AF3 [8] & AF3 [16] & Boltz-2 [8] & Boltz-2 [16] \\
\midrule
\endhead
\midrule
\multicolumn{6}{r}{\emph{Continued on next page}}\\
\endfoot
\bottomrule
\endlastfoot
cLDDT $\uparrow$ & \shortstack{-.0056\\{}[-.0071, -.0042]} & \shortstack{+.0020\\{}[.0012, .0028]} & \shortstack{+.0019\\{}[.0013, .0026]} & \shortstack{-.0007\\{}[-.0020, .0005]} & \shortstack{+.0034\\{}[.0025, .0044]} \\
bbLDDT $\uparrow$ & \shortstack{-.0008\\{}[-.0017, .0002]} & \shortstack{+.0014\\{}[.0007, .0021]} & \shortstack{+.0015\\{}[.0009, .0021]} & \shortstack{+.0012\\{}[.0001, .0022]} & \shortstack{+.0025\\{}[.0016, .0034]} \\
TM $\uparrow$ & \shortstack{-.0009\\{}[-.0026, .0007]} & \shortstack{-.0010\\{}[-.0027, .0005]} & \shortstack{-.0004\\{}[-.0020, .0011]} & \shortstack{+.0016\\{}[-.0003, .0034]} & \shortstack{+.0024\\{}[.0006, .0042]} \\
DockQ $\uparrow$ & \shortstack{-.0013\\{}[-.0037, .0012]} & \shortstack{+.0009\\{}[-.0017, .0033]} & \shortstack{-.0031\\{}[-.0057, -.0006]} & \shortstack{-.0031\\{}[-.0060, -.0002]} & \shortstack{+.0003\\{}[-.0024, .0030]} \\
QS $\uparrow$ & \shortstack{+.0021\\{}[-.0003, .0045]} & \shortstack{-.0016\\{}[-.0041, .0009]} & \shortstack{.0000\\{}[-.0024, .0023]} & \shortstack{-.0027\\{}[-.0059, .0003]} & \shortstack{-.0019\\{}[-.0049, .0010]} \\
ICS $\uparrow$ & \shortstack{+.0026\\{}[.0003, .0048]} & \shortstack{.0000\\{}[-.0023, .0023]} & \shortstack{-.0016\\{}[-.0037, .0006]} & \shortstack{-.0027\\{}[-.0056, -.0001]} & \shortstack{-.0020\\{}[-.0046, .0005]} \\
IPS $\uparrow$ & \shortstack{+.0014\\{}[-.0009, .0037]} & \shortstack{+.0014\\{}[-.0008, .0036]} & \shortstack{+.0010\\{}[-.0011, .0031]} & \shortstack{-.0001\\{}[-.0025, .0022]} & \shortstack{+.0004\\{}[-.0019, .0027]} \\
lDDT-PLI $\uparrow$ & \shortstack{-.0065\\{}[-.0120, -.0013]} & \shortstack{-.0016\\{}[-.0064, .0030]} & \shortstack{.0000\\{}[-.0054, .0050]} & \shortstack{-.0051\\{}[-.0098, -.0003]} & \shortstack{-.0023\\{}[-.0069, .0022]} \\
RMSD (\AA) $\downarrow$ & \shortstack{-0.19\\{}[-0.42, 0.04]} & \shortstack{-0.17\\{}[-0.39, 0.05]} & \shortstack{-0.12\\{}[-0.34, 0.10]} & \shortstack{-0.26\\{}[-0.55, 0.04]} & \shortstack{-0.22\\{}[-0.52, 0.07]} \\
Assigned $<2$\AA{} $\uparrow$ & \shortstack{-0.3\\{}[-1.4, 0.9]} & \shortstack{+0.0\\{}[-1.0, 1.0]} & \shortstack{+0.1\\{}[-1.0, 1.1]} & \shortstack{-0.5\\{}[-1.3, 0.3]} & \shortstack{+0.1\\{}[-0.6, 0.7]} \\
PB-valid $\uparrow$ & \shortstack{-10.4\\{}[-13.0, -7.7]} & \shortstack{-2.0\\{}[-4.4, 0.2]} & \shortstack{+0.8\\{}[-1.3, 2.9]} & \shortstack{-8.9\\{}[-10.9, -6.9]} & \shortstack{-5.0\\{}[-6.8, -3.4]} \\
Primary $<2$\AA{} $\uparrow$ & \shortstack{-0.4\\{}[-1.9, 1.1]} & \shortstack{+0.1\\{}[-1.2, 1.4]} & \shortstack{+0.3\\{}[-0.9, 1.7]} & \shortstack{-0.6\\{}[-1.9, 0.8]} & \shortstack{-0.2\\{}[-1.4, 1.0]} \\
Success $\uparrow$ & \shortstack{-4.3\\{}[-6.4, -2.3]} & \shortstack{+1.1\\{}[-0.6, 2.7]} & \shortstack{+1.2\\{}[-0.3, 2.6]} & \shortstack{-5.9\\{}[-7.6, -4.4]} & \shortstack{-2.6\\{}[-3.9, -1.5]} \\
RF-valid $\uparrow$ & \shortstack{-4.6\\{}[-6.1, -3.1]} & \shortstack{-1.0\\{}[-2.4, 0.5]} & \shortstack{+0.8\\{}[-0.7, 2.3]} & \shortstack{-4.8\\{}[-5.9, -3.9]} & \shortstack{-4.4\\{}[-5.3, -3.4]} \\
Bonds $\downarrow$ & \shortstack{+12.79\\{}[7.69, 19.71]} & \shortstack{+0.73\\{}[0.23, 1.50]} & \shortstack{+0.04\\{}[0.02, 0.06]} & \shortstack{+3.94\\{}[2.46, 6.10]} & \shortstack{+0.88\\{}[0.36, 1.69]} \\
Angles $\downarrow$ & \shortstack{+3.92\\{}[2.45, 6.22]} & \shortstack{+0.15\\{}[0.07, 0.25]} & \shortstack{-0.01\\{}[-0.02, 0.01]} & \shortstack{+2.34\\{}[1.32, 3.86]} & \shortstack{+0.52\\{}[0.17, 1.02]} \\
Clashes $\downarrow$ & \shortstack{+23.52\\{}[10.49, 43.90]} & \shortstack{+3.25\\{}[1.07, 6.65]} & \shortstack{+1.06\\{}[-0.02, 3.11]} & \shortstack{+3.48\\{}[2.17, 5.25]} & \shortstack{+0.81\\{}[0.26, 1.53]} \\
Any violation $\downarrow$ & \shortstack{+47.6\\{}[45.5, 49.7]} & \shortstack{+1.5\\{}[0.4, 2.6]} & \shortstack{-3.0\\{}[-4.0, -2.1]} & \shortstack{+11.6\\{}[9.7, 13.5]} & \shortstack{-3.6\\{}[-5.3, -1.9]} \\
\end{longtable}
\noindent Each comparison uses its own metric-specific teacher--student intersection and joint PDB-cluster resampling. These supports can differ from the multi-model tables. Differences in rates are percentage points. Intervals are pointwise and do not measure variation across training runs.\par
\endgroup

\section{Numerical analysis of the molecular sampler}
\label{app:numerical}
We test whether the schedule improvement survives changes in minimum
noise, then examine how individual sampling intervals affect the final
structure. The schedule sweeps use the Boltz-2 benchmark; the interval
diagnostics use selected targets with paired candidates.

\subsection{Controlled sampler and minimum-noise sensitivity}
The AF3 field centers valid atomic coordinates using
$D_c(x,\sigma;c)=P D_\phi(Px,\sigma;c)$, where $P$ subtracts the
valid-atom centroid. Conditioning, masks, initialization, and the coordinate
frame are fixed. These deterministic teacher experiments use scale-one
Euler updates without churn, random frame augmentation, or steering.

The budget and minimum-noise sweeps cover 44 configurations with five
candidates per target on the Boltz-2 benchmark ($\sim$2,300 targets).
Accuracy uses the complete-case intersection described in
Appendix~\ref{app:datasets}. We average candidates within each target,
then weight targets equally. All reported lDDT includes stereochemical checks.

\begin{figure}[!htbp]
\centering\includegraphics[width=.86\linewidth]{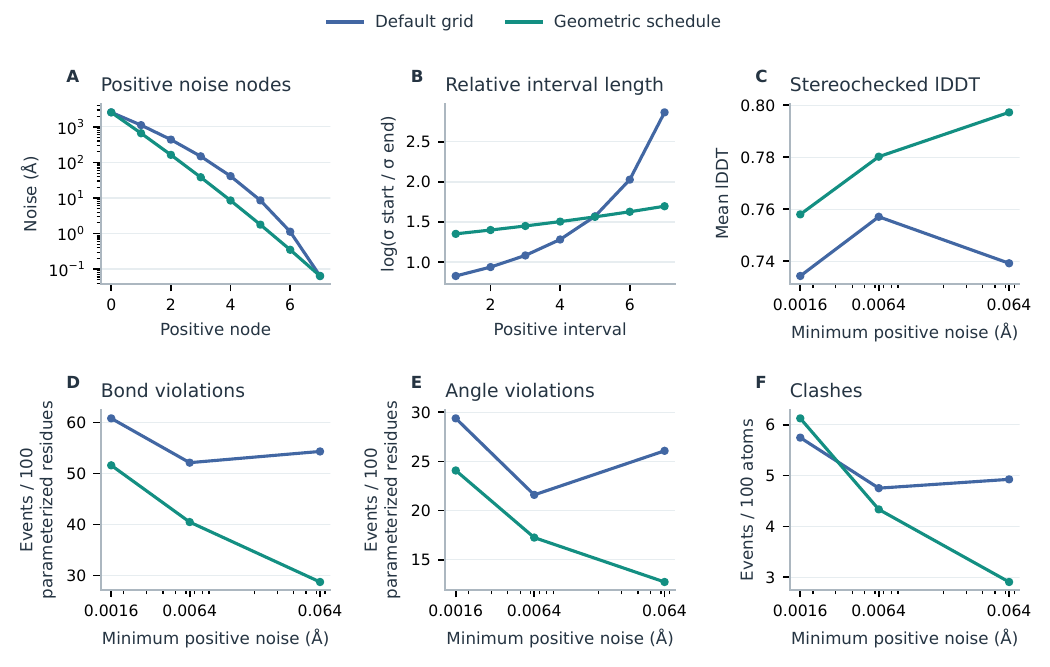}
\caption{\textbf{Noise grids and minimum-noise sensitivity.}
A--B: eight positive nodes and seven relative intervals, with maximum
noise 2,560~\AA{} and minimum 0.064~\AA{}; terminal zero is separate.
C--F: accuracy and geometry at eight evaluations as the minimum noise
changes. Bands are target-bootstrap intervals. Bond and angle rates are
per 100 parameterized residues; clashes are per 100 atoms.}
\label{fig:nodes-minima}
\end{figure}
Changing the minimum noise alone does not reproduce the geometric
schedule's combination of accuracy and geometry
(Figure~\ref{fig:nodes-minima}). The schedule advantage also depends on
the step budget, as shown in Figure~\ref{fig:budget}.

\subsection{Same-input local defects}
Each selected diagnostic target contributes two paired candidates.
We evaluate intervals starting from both coarse and refined trajectories.
From each starting state, Euler, Heun, and the refined calculation use
the same input and coordinate frame. Local defects use component RMSE,
$\|e\|_F/\sqrt{3N}$ for $N$ valid atoms; endpoint displacement uses
unaligned atom RMSD, $\|e\|_F/\sqrt{N}$.

\begin{figure}[!htbp]
\centering\includegraphics[width=.86\linewidth]{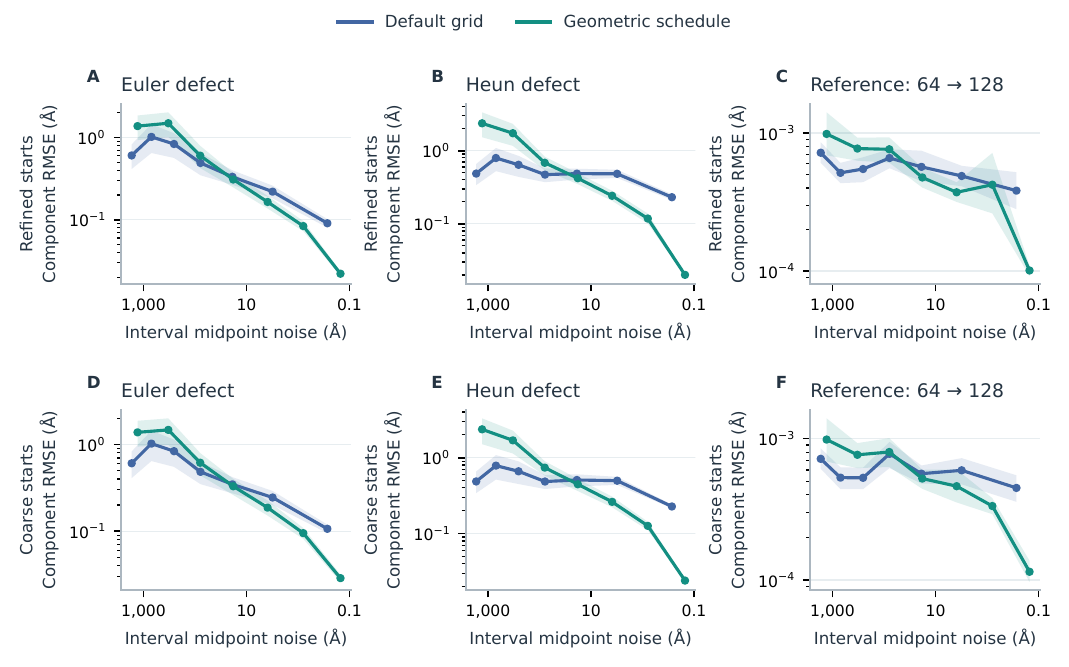}
\caption{\textbf{Local defects and reference sensitivity.}
Top: refined starting states. Bottom: coarse starting states.
Columns show single-step Euler and Heun defects against a
128-subdivision reference, followed by the 64-to-128-subdivision
reference difference. Curves average candidates within targets; bands
resample targets. The horizontal axis gives interval midpoints in noise.}
\label{fig:local-full}
\end{figure}
The reference-refinement differences are much smaller than the
single-step defects for both starting-state constructions
(Figure~\ref{fig:local-full}). This checks sensitivity to the numerical
reference without treating it as an exact solution.

\subsection{Single-interval interventions and final geometry}
For Equation~\ref{eq:effect}, the control follows the all-refined path.
The branch shares its prefix, substitutes one coarse Euler step, and
then follows the same refined continuation. We score the final structures.

\begin{figure}[!htbp]
\centering\includegraphics[width=.86\linewidth]{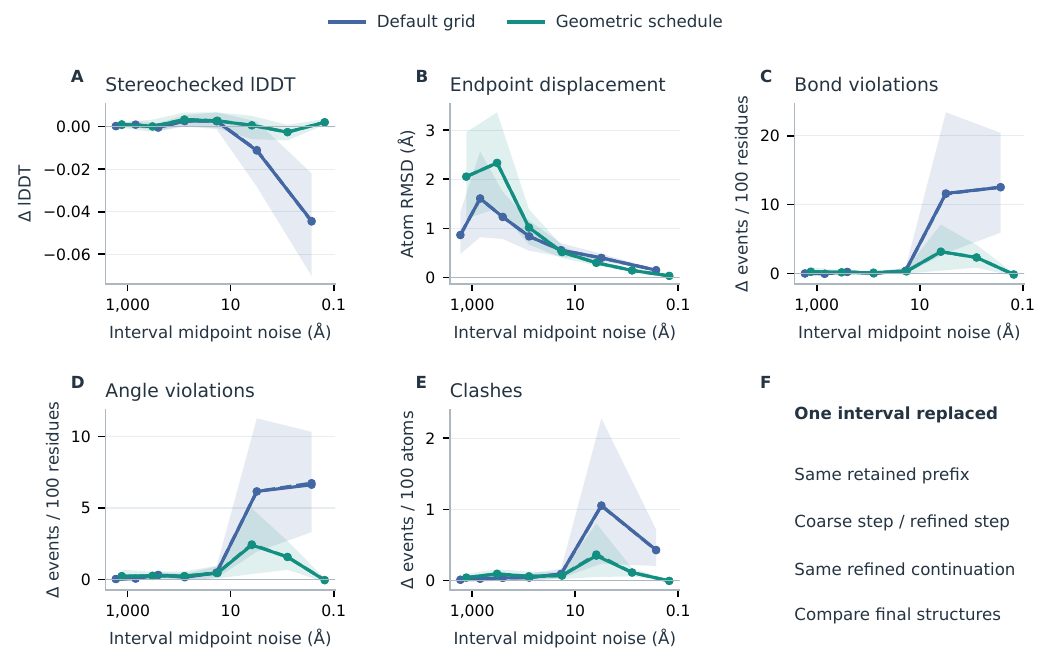}
\caption{\textbf{How one coarse interval affects the final structure.}
Solid and dashed curves use 200 and 100 subdivisions in the refined
calculation. A and C--E show branch minus control; B shows unaligned
endpoint atom RMSD. Negative accuracy changes and positive violation
changes are worse. Bond and angle rates are per 100 parameterized
residues; clashes are per 100 atoms. Comparisons pair targets and
candidates; bands resample targets.}
\label{fig:intervention-full}
\end{figure}
For the default grid, the last positive interval causes the strongest
mean loss of stereochemically checked lDDT, although an earlier interval
produces a larger endpoint displacement (Figure~\ref{fig:intervention-full}).
Bond and angle violations also rise at low noise. The geometric grid
changes this damage profile, showing why coordinate displacement alone
is insufficient to judge the quality of an update. The grids cover
different noise windows, so equal interval indices are not directly
matched. Nor can the individual effects be added: an earlier change
would alter the starting states of later intervals.

\end{document}